\documentclass{article}

\usepackage[final,main]{arxiv_2026}

\usepackage[utf8]{inputenc} 
\usepackage[T1]{fontenc}    
\usepackage{hyperref}       
\usepackage{url}            
\usepackage{booktabs}       
\usepackage{amsfonts}       
\usepackage{nicefrac}       
\usepackage{microtype}      
\usepackage{xcolor}         
\usepackage{amsmath,amssymb}
\usepackage{algorithm}
\usepackage{algorithmic}
\usepackage{multirow}
\usepackage{graphicx}
\usepackage{tikz}
\usepackage[table]{xcolor}
\usetikzlibrary{decorations.pathreplacing,calc}
\usepackage{enumitem}

\usetikzlibrary{positioning,fit,arrows.meta}
\usepackage{bm}

\title{Performance-Driven Policy Optimization for Speculative Decoding with Adaptive Windowing}

\author{
Jie Jiang$^{1}$\thanks{These authors contributed equally to this work.} \quad
Xing Sun$^{2}$\footnotemark[1] \quad
Ruotian Chen$^{1}$ \quad
Jianan Su$^{1}$\thanks{Corresponding authors.} \quad
Kaixin Shen$^{1}$\footnotemark[2] \\[0.5em]
$^{1}$Ant International, Ant Group \\
$^{2}$Fliggy, Alibaba Group \\[0.5em]
\texttt{\{youzhen.jj, ruotian.crt, jianan.sja, kaxin.skx\}@ant-intl.com} \\
\texttt{ruoxing.sx@alibaba-inc.com}
}

\begin{document}

\maketitle

\begin{abstract}
  Speculative decoding accelerates LLM inference by having a lightweight draft model propose speculative windows of candidate tokens for parallel verification by a larger target model. In practice, speculative efficiency is often bottlenecked by hard-to-draft positions, where an early mismatch truncates the accepted prefix and invalidates the rest of the speculative window. Most learning-based drafters are still optimized with token-level supervised objectives, even though speculative utility is inherently window-level and prefix-sensitive. We propose \textbf{PPOW} (\textbf{P}erformance-Driven \textbf{P}olicy \textbf{O}ptimization with Adaptive \textbf{W}indowing), a reinforcement learning framework that shifts drafter optimization from token-level imitation to window-level optimization. PPOW combines a Cost-Aware Speedup Reward, a Distribution-Based Proximity Reward, and Adaptive Divergence-Aware Windowing, which prioritizes informative windows with high confidence-weighted draft--target divergence. PPOW achieves average acceptance lengths of 6.29--6.52 and speedups of 3.39--4.36$\times$ across multiple model families and benchmarks under a unified decoding protocol. These results show that performance-driven window-level optimization is a practical approach to improving speculative decoding efficiency.
\end{abstract}

\section{Introduction}
Speculative decoding~\cite{leviathan2023fast, chen2023accelerating} accelerates Large Language Model (LLM) generation while preserving the output distribution of the target model. At each speculative step, a lightweight draft model (drafter) proposes a \emph{speculative window} of candidate tokens, which the target model then verifies in parallel. In practice, the realized speedup is often limited by hard-to-draft positions. Even when most tokens within a speculative window are well modeled by the drafter, a single early mismatch can truncate the accepted prefix and invalidate the rest of the window, limiting overall throughput.

This behavior exposes a mismatch between the inference-time goal of speculative decoding and the training objectives commonly used for drafters. Recent methods, including MEDUSA~\cite{cai2024medusa}, Hydra~\cite{ankner2024hydra}, EAGLE~\cite{li2024eagle, li2024eagle2, li2025eagle3}, HASS~\cite{zhang2025learning}, and GRIFFIN~\cite{hu2025griffin}, improve drafter quality through architectural design and supervised training, while distillation-based approaches~\cite{zhou2024distillspec, liu2024online} further improve distributional alignment with the target model. However, these methods are still primarily based on token-level supervised objectives. Such objectives improve local next-token prediction, but speculative utility is inherently \emph{window-level} and \emph{prefix-sensitive}: once verification is truncated at an early position, the remaining drafted tokens in the speculative window are invalidated. As a result, improvements in token-level imitation may not consistently lead to longer accepted prefixes or higher speculative efficiency, especially when end-to-end performance is governed by a few acceptance bottlenecks.

We propose \textbf{PPOW} (\textbf{P}erformance-Driven \textbf{P}olicy \textbf{O}ptimization with Adaptive \textbf{W}indowing), a reinforcement learning framework that shifts drafter training from token-level supervision to window-level optimization over speculative windows. Figure~\ref{fig:ppow_frame} provides an overview of PPOW. This window-level formulation aligns optimization with the prefix-sensitive structure induced by speculative verification. Concretely, PPOW combines a \emph{Cost-Aware Speedup Reward}, which encourages longer accepted prefixes while accounting for relative drafter cost, with a \emph{Distribution-Based Proximity Reward}, an auxiliary signal that preserves partial credit for speculative windows that remain close to the target model's preferences even when the accepted prefix is truncated early. 

Although PPOW optimizes with window-level rewards, not all speculative windows are equally informative during training. Treating all windows uniformly can disperse optimization effort across non-bottleneck windows, limiting focus on the acceptance bottlenecks most critical to speculative efficiency. PPOW therefore introduces \emph{Adaptive Divergence-Aware Windowing} (ADAW), which prioritizes windows with large confidence-weighted draft--target divergence. This criterion is grounded in our analysis of potential acceptance bottlenecks in Appendix~\ref{app:proofs}, and our experiments show that prioritizing such windows improves speculative performance.

Our contributions are summarized as follows:
\begin{itemize}[leftmargin=3em, nosep, parsep=1.8pt]
\item \textbf{A Window-Level RL Framework for Drafter Optimization.} We formulate drafter training for speculative decoding as a reinforcement learning problem over speculative windows, which better matches inference-time acceptance behavior.
\item \textbf{Performance-Driven Reward Design.} PPOW combines a Cost-Aware Speedup Reward with an auxiliary Distribution-Based Proximity Reward to better align training with speculative decoding; the latter provides additional credit for speculative windows that remain close to the target distribution despite early truncation of the accepted prefix, as illustrated in Figure~\ref{fig:token_mismatch}.
\item \textbf{Adaptive Divergence-Aware Windowing.} We introduce a window prioritization strategy based on confidence-weighted draft--target divergence that focuses training on more informative windows associated with acceptance bottlenecks. Our analysis and experiments show that prioritizing such windows improves speculative performance.
\end{itemize}

\begin{figure}[t]
\centering
\begingroup
\definecolor{tmInk}{HTML}{1D1D1F}
\definecolor{tmFlow}{HTML}{6E6E73}
\definecolor{tmBorder}{HTML}{D6D6DC}
\definecolor{tmContext}{HTML}{F6F4F1}
\definecolor{tmDraft}{HTML}{FFE7D1}
\definecolor{tmDraftBorder}{HTML}{F4DDC7}
\definecolor{tmTarget}{HTML}{F0F0F3}
\definecolor{tmReward}{HTML}{EAF5FF}
\definecolor{tmBlue}{HTML}{D0E7FF}
\definecolor{tmMint}{HTML}{DCF5F1}
\definecolor{tmGreen}{HTML}{ADE6B8}
\definecolor{tmPink}{HTML}{F6DEDF}
\definecolor{tmCheck}{HTML}{299E63}
\definecolor{tmCross}{HTML}{CF5B5E}
\resizebox{0.98\linewidth}{!}{
\begin{tikzpicture}[
    every node/.style={font=\rmfamily\normalsize, text=tmInk},
    line cap=round, line join=round,
    input_box/.style={
    draw=tmBorder, fill=tmContext, rounded corners=5pt,
    minimum width=5.5cm,
    minimum height=0.62cm,
    inner sep=5pt,
    font=\rmfamily\normalsize
    },
    token_d/.style={
        draw=tmDraftBorder, fill=tmDraft, rounded corners=3pt,
        minimum width=0.82cm, minimum height=0.42cm,
        thick, font=\rmfamily\small
    },
    token_t/.style={
        draw=tmBorder, fill=tmTarget!50!white, rounded corners=3pt,
        minimum width=0.82cm, minimum height=0.42cm,
        dashed, font=\rmfamily\small
    },
    token_acc/.style={
        draw=tmGreen, fill=tmMint, rounded corners=3pt,
        minimum width=0.82cm, minimum height=0.42cm,
        thick, font=\rmfamily\small
    },
    token_rej/.style={
        draw=tmCross!45!white, fill=tmPink, rounded corners=3pt,
        minimum width=0.82cm, minimum height=0.42cm,
        thick, font=\rmfamily\small
    },
    label_node/.style={font=\rmfamily\bfseries\small, text=tmFlow},
    panel_title/.style={font=\rmfamily\bfseries\small},
    outcome_box/.style={
        draw=tmBlue, fill=tmReward, rounded corners=6pt,
        align=center,
        text width=3.9cm,
        minimum height=1.15cm,
        inner sep=4pt,
        font=\rmfamily\small
    },
    formula_box/.style={
        draw=tmBorder, rounded corners=6pt, inner sep=4pt,
        align=center, text width=5.0cm, font=\rmfamily\small, fill=white
    }
]

\node[panel_title] at (-5.3, 0.65) {(a) Cost-Aware Speedup Reward};

\node[input_box] (inputA) at (-5.3, -0.05) {\textbf{Context:} The answer is};

\node[text=tmCheck, font=\rmfamily\small] at (-6.0, -0.62) {accepted prefix: $k=3$};

\node[token_acc] (ad1) at (-7.4, -1.15) {8};
\node[token_acc] (ad2) at (-6.35, -1.15) {/};
\node[token_acc] (ad3) at (-5.3, -1.15) {9};
\node[token_rej] (ad4) at (-4.25, -1.15) {.};
\node[token_rej] (ad5) at (-3.2, -1.15) {EOS};

\node[anchor=east, label_node] at (-8.05, -1.15) {Draft};

\node[token_t] (at1) at (-7.4, -2.02) {8};
\node[token_t] (at2) at (-6.35, -2.02) {/};
\node[token_t] (at3) at (-5.3, -2.02) {9};
\node[token_t] (at4) at (-4.25, -2.02) {,};
\node[token_t] (at5) at (-3.2, -2.02) {then};

\node[anchor=east, label_node] at (-8.05, -2.02) {Target};

\draw[decorate,decoration={brace,amplitude=4pt}, thick, tmCheck]
    (-7.8,-0.90) -- (-4.8,-0.90);

\node[outcome_box] (rewardA) at (-5.3, -3.45) {
    \shortstack[c]{
        \textbf{Cost-Aware Speedup}\\[1.5pt]
        $R_{\text{speedup}} = k \,/\, (k\gamma + 1)$
    }
};

\draw[->, >=stealth, color=tmCheck, thick] (at3.south) -- (rewardA.north);

\node[panel_title] at (5.3, 0.65) {(b) Distribution-Based Proximity Reward};

\node[input_box] (inputB) at (5.3, -0.05) {\textbf{Context:} The math result is};

\node[text=tmCross, font=\rmfamily\small] at (5.3, -0.62) {early truncation $\Rightarrow k=0$};

\node[token_d] (bd1) at (3.2, -1.15) {\$};
\node[token_d] (bd2) at (4.25, -1.15) {$\backslash$frac};
\node[token_d] (bd3) at (5.3, -1.15) {\{8\}};
\node[token_d] (bd4) at (6.35, -1.15) {\{9\}};
\node[token_d] (bd5) at (7.4, -1.15) {\$};

\node[anchor=east, label_node] at (2.55, -1.15) {Draft};

\node[token_t] (bt1) at (3.2, -2.02) {**};
\node[token_t] (bt2) at (4.25, -2.02) {8};
\node[token_t] (bt3) at (5.3, -2.02) {/};
\node[token_t] (bt4) at (6.35, -2.02) {9};
\node[token_t] (bt5) at (7.4, -2.02) {**};

\node[anchor=east, label_node] at (2.55, -2.02) {Target};

\node[outcome_box, draw=tmCross!45!white, fill=tmPink!65!white] (rewardB1) at (3.0, -3.45) {
    \shortstack[c]{
        \textbf{Acceptance Feedback}\\[1.5pt]
        $k=0,\; R_{\text{speedup}}=0$
    }
};

\node[outcome_box] (rewardB2) at (7.6, -3.45) {
    \shortstack[c]{
        \textbf{Distribution-Based Proximity}\\[1.5pt]
        $R_{\text{dist}}=\eta,\; \text{ if } \Delta<\epsilon$
    }
};

\draw[->, >=stealth, color=tmCross, thick] (bt2.south) -- (rewardB1.north);
\draw[->, >=stealth, color=tmCheck, thick] (bt4.south) -- (rewardB2.north);

\end{tikzpicture}
}
\endgroup
\caption{\textbf{PPOW uses a Cost-Aware Speedup Reward together with a Distribution-Based Proximity Reward.}
(a) The Cost-Aware Speedup Reward increases with accepted prefix length and directly encourages speculative decoding efficiency.
(b) When verification is truncated early, resulting in $k=0$, the Distribution-Based Proximity Reward still provides auxiliary credit if the speculative window remains close to target preferences.}
\label{fig:token_mismatch}
\end{figure}

\begin{figure*}[t]
\centering
\begingroup
\definecolor{ppowInk}{HTML}{1D1D1F}
\definecolor{ppowFlow}{HTML}{6E6E73}
\definecolor{ppowBorder}{HTML}{D6D6DC}
\definecolor{ppowHighStage}{HTML}{F8FAFC}
\definecolor{ppowAdaptiveStage}{HTML}{F8F7FC}
\definecolor{ppowTrainingStage}{HTML}{FBFAF8}
\definecolor{ppowHighHeader}{HTML}{EAF4FB}
\definecolor{ppowAdaptiveHeader}{HTML}{EEEAFE}
\definecolor{ppowTrainingHeader}{HTML}{F8EFE6}
\definecolor{ppowTarget}{HTML}{DCEEFF}
\definecolor{ppowDraft}{HTML}{FFE7D1}
\definecolor{ppowAdaptive}{HTML}{F4DDC7}
\definecolor{ppowReward}{HTML}{EAF5FF}
\definecolor{ppowColumn}{HTML}{F0F0F3}
\definecolor{ppowInputsPanel}{HTML}{F6F4F1}
\definecolor{ppowInputsHeader}{HTML}{EEE8DF}
\definecolor{ppowMint}{HTML}{DCF5F1}
\definecolor{ppowPink}{HTML}{F6DEDF}
\definecolor{ppowGreen}{HTML}{ADE6B8}
\definecolor{ppowBlue}{HTML}{D0E7FF}
\definecolor{ppowPurple}{HTML}{E8DDFF}
\definecolor{ppowSelected}{HTML}{DE6567}
\definecolor{ppowCheck}{HTML}{299E63}
\definecolor{ppowCross}{HTML}{CF5B5E}

\newcommand{\ppowtext}[4]{%
  \node[font=\rmfamily\fontsize{#3bp}{#3bp}\selectfont] at (#1,#2) {#4};}
\newcommand{\ppowfit}[6]{%
  \ppowtext{#1}{#2}{#3}{\scalebox{#4}[#5]{#6}}}

\newcommand{\ppowHeaderTop}{70}
\newcommand{\ppowHeaderBottom}{68}
\newcommand{\ppowHeaderTextYOffset}{38}

\newcommand{\ppowheading}[3]{%
  \node[
    anchor=west,
    font=\rmfamily\bfseries\fontsize{11bp}{11bp}\selectfont,
    text=ppowInk
  ] at (#1,#2) {#3};}

\newcommand{\ppowstage}[7]{%
  \begin{scope}
    \clip[rounded corners=16bp] (#1,#2) rectangle (#3,#4);
    \fill[#5] (#1,#2) rectangle (#3,#4);
    \fill[#6,rounded corners=12bp]
      (#1,#2-\ppowHeaderBottom) rectangle (#3,#2+\ppowHeaderTop);
  \end{scope}
  \draw[ppowFrame] (#1,#2) rectangle (#3,#4);
  \ppowheading{#1+41}{#2+\ppowHeaderTextYOffset}{#7}}

\newcommand{\ppowtick}[2]{%
  \draw[ppowCheck,line width=1.1bp,line cap=round,line join=round]
    (#1-7,#2) -- (#1-2,#2+5.5) -- (#1+8,#2-7);}
\newcommand{\ppowcross}[2]{%
  \draw[ppowCross,line width=1.0bp,line cap=round]
    (#1-6,#2-6) -- (#1+6,#2+6)
    (#1-6,#2+6) -- (#1+6,#2-6);}
\newcommand{\ppowTopExtension}{100}

\resizebox{\textwidth}{!}{%
\begin{tikzpicture}[
  x=0.2bp,y=-0.2bp,
  every node/.style={inner sep=0pt,outer sep=0pt,text=ppowInk},
  draw=ppowBorder,
  line width=0.6bp,line cap=round,line join=round,
  ppowFrame/.style={rounded corners=16bp,line width=0.7bp},
  ppowArrow/.style={draw=ppowFlow,-{Stealth[length=3.3bp,width=3bp]},
                    line width=0.8bp},
  ppowPill/.style={fill=white,rounded corners=9bp,line width=0.45bp},
  ppowModel/.style={draw=none,rounded corners=14bp},
  ppowSmall/.style={fill=white,rounded corners=7bp,line width=0.45bp}
]
\path[use as bounding box] (0,-\ppowTopExtension) rectangle (3264,900);
\clip (0,-\ppowTopExtension) rectangle (3264,900);
\fill[white] (0,-\ppowTopExtension) rectangle (3264,900);

\ppowstage{148}{2-\ppowTopExtension}{1270}{293}{ppowHighStage}{ppowHighHeader}{High-Criticality Window}
\ppowstage{148}{330}{1270}{899}{ppowAdaptiveStage}{ppowAdaptiveHeader}{Adaptive Windowing Stage}
\ppowstage{1564}{2-\ppowTopExtension}{3263}{899}{ppowTrainingStage}{ppowTrainingHeader}{Training Stage}

\fill[ppowMint,rounded corners=4.0bp] (334,149) rectangle (595,199);
\fill[ppowPink,rounded corners=4.0bp] (718,149) rectangle (1069,199);
\ppowfit{721.5}{164.5}{12}{0.988}{1}{$(\boldsymbol{x},y_1,\ldots,y_K,\ldots,y_t,\ldots,y_{t+K-1},\ldots)$}
\ppowtext{463}{246}{11}{$s_1$}
\ppowfit{659.33}{246.6}{11}{1.167}{1.4}{$\cdots$}
\ppowfit{846.53}{245.53}{11}{1.028}{1.031}{$s_i$}

\draw[ppowArrow] (717,87) -- (835,144);
\node[
  anchor=center,
  font=\rmfamily\fontsize{10.5bp}{10.5bp}\selectfont,
  text=ppowInk,fill=ppowPink,
  rounded corners=4.0bp,
  inner xsep=5bp,
  inner ysep=2.5bp
] at (709,50) {hard-to-draft};

\begin{scope}[shift={(0,14)}]
\draw[ppowPill] (2,595) rectangle (128,684);
\ppowfit{63.5}{637}{15}{0.93}{1}{$\boldsymbol{x}$}
\draw[ppowModel,fill=ppowTarget] (187,576) rectangle (423,702);
\ppowtext{305}{614}{9.5}{Target}
\ppowtext{305}{663}{9.5}{Model}
\draw[ppowPill] (474,483) rectangle (600,571);
\ppowfit{537.5}{530}{14}{0.865}{0.87}{$\boldsymbol{y}$}
\draw[ppowPill] (474,595) rectangle (600,684);
\ppowfit{541.5}{637}{10.5}{0.849}{0.969}{$\mathbf{H}^{t_x}_{1:n_1}$}
\draw[ppowPill] (474,707) rectangle (600,795);
\ppowfit{537.5}{751}{10.5}{0.831}{0.914}{$\mathbf{H}^{t_y}_{1:n_2}$}
\draw[ppowArrow] (134,638) -- (181,638);
\draw[ppowArrow] (429,638) -- (467,638);
\draw[ppowArrow] (606,638) -- (648,638);

\draw[ppowFrame,draw=ppowGreen] (654,450) rectangle (1056,827);
\fill[ppowAdaptive,rounded corners=12bp]
  (672,475) rectangle (1038,630);
\ppowtext{855}{503}{9.5}{Adaptive}
\ppowtext{855}{552}{9.5}{Divergence-Aware}
\ppowtext{855}{600}{9.5}{Windowing}
\draw[ppowModel,fill=ppowDraft] (740,667) rectangle (974,794);
\ppowtext{857}{709}{9.5}{Draft}
\ppowtext{857}{757}{9.5}{Model}
\draw[ppowArrow] (1063,638) -- (1101,638);

\fill[ppowColumn,rounded corners=12bp] (1110,444) rectangle (1254,826);
\draw[ppowSmall] (1130,469) rectangle (1233,538);
\draw[ppowSmall] (1130,557) rectangle (1233,625);
\draw[ppowSmall,draw=ppowSelected,line width=1bp] (1130,646) rectangle (1233,715);
\draw[ppowSmall] (1130,735) rectangle (1233,804);
\ppowfit{1181.93}{505.46}{12}{.953}{.914}{$s_1$}
\foreach \ppowDotY in {574.5,592.5,611}
  \fill[black] (1182,\ppowDotY) ellipse[x radius=1.5,y radius=2.5];
\ppowfit{1181.97}{680.46}{12}{.946}{.914}{$s_i$}
\foreach \ppowDotY in {754,772,790.5}
  \fill[black] (1182,\ppowDotY) ellipse[x radius=1.5,y radius=2.5];
\end{scope}

\begin{scope}
  \clip[rounded corners=12bp] (1333,265) rectangle (1525,610);
  \fill[ppowInputsPanel] (1333,265) rectangle (1525,610);
  \fill[ppowInputsHeader,rounded corners=10bp] (1333,205) rectangle (1525,327);
\end{scope}
\draw[rounded corners=12bp,line width=0.6bp] (1333,265) rectangle (1525,610);
\node[font=\rmfamily\bfseries\fontsize{9bp}{9bp}\selectfont] at (1429,298) {Inputs};
\draw[ppowPill] (1344,351) rectangle (1513,439);
\draw[ppowPill] (1344,499) rectangle (1513,588);
\ppowfit{1433.5}{399.5}{13}{0.918}{1.024}{$\boldsymbol{x},\boldsymbol{y}_{<t}$}
\ppowfit{1433.5}{543.5}{11}{0.825}{0.932}{$\mathbf{H}^{t_{x,y}}_{1:n_1+t}$}
\draw[ppowArrow] (1241,694) .. controls (1328,694) and (1243,466) .. (1327,466);
\draw[ppowArrow] (1531,466) -- (1590,466);

\ppowfit{1942}{245.5}{9}{0.987}{0.923}{Candidate Group}
\ppowfit{2238.5}{318.5}{11}{0.922}{0.878}{$KL$}
\ppowfit{3130.5}{82}{11}{0.987}{0.918}{Advantage}

\draw[ppowArrow] (2208,398)
  .. controls (2146.64,115.68) and (1678.89,125.71) .. (1710,396);
\draw[ppowArrow] (3150,269)
  .. controls (3056.71,-151.35) and (1642.13,-17.40) .. (1710,396);

\draw[ppowModel,fill=ppowDraft] (1596,403) rectangle (1831,530);
\ppowtext{1713.5}{445}{9.5}{Draft}
\ppowtext{1713.5}{489}{9.5}{Model}
\draw[ppowArrow] (1839,466) -- (1880,466);
\fill[ppowColumn,rounded corners=12bp] (1886,270) rectangle (2031,663);
\draw[ppowSmall] (1907,293) rectangle (2010,364);
\draw[ppowSmall] (1907,386) rectangle (2010,456);
\draw[ppowSmall] (1907,478) rectangle (2010,548);
\draw[ppowSmall] (1907,570) rectangle (2010,640);
\ppowfit{1962.5}{327}{12}{0.71}{0.8}{$\hat{\boldsymbol{y}}_1$}
\ppowfit{1962.5}{420}{12}{0.71}{0.8}{$\hat{\boldsymbol{y}}_2$}
\foreach \ppowDotY in {497,515.5,534}
  \fill[black] (1958,\ppowDotY) ellipse[x radius=1.5,y radius=2.5];
\ppowfit{1962.5}{603}{12}{0.71}{0.8}{$\hat{\boldsymbol{y}}_G$}
\draw[ppowArrow] (2037,466) -- (2086,466);
\draw[ppowModel,fill=ppowTarget] (2092,404) rectangle (2325,530);
\ppowtext{2208.5}{440}{9.5}{Target}
\ppowtext{2208.5}{485}{9.5}{Model}

\draw[ppowPill] (1867,714) rectangle (2048,798);
\ppowfit{1957.5}{755}{12}{0.80}{0.85}{$\boldsymbol{y}_{t:t+K-1}$}
\draw[ppowArrow] (2048,755)
  .. controls (2157,745) and (2166,594) .. (2095,500);
\draw[ppowArrow] (2331,467) -- (2373,467);

\draw[ppowFrame,draw=ppowBlue] (2380,48) rectangle (2818,451);
\draw[ppowFrame,draw=ppowPurple] (2380,484) rectangle (2818,877);
\draw[ppowModel,fill=ppowReward] (2418,305) rectangle (2785,438);
\ppowtext{2601.5}{340}{8.5}{Cost-Aware}
\ppowtext{2601.5}{386}{8.5}{Speedup Reward}
\draw[ppowModel,fill=ppowReward] (2418,496) rectangle (2785,630);
\ppowtext{2601.5}{541}{8.5}{Distribution-Based}
\ppowtext{2601.5}{586}{8.5}{Proximity Reward}

\foreach \row/\leftword/\rightword in {
  0/Life/Life,1/is/is,2/a/roll,3/box/of,4/of/film}{
  \pgfmathsetmacro{\ppowY}{74+48*\row}
  \ppowtext{2511}{\ppowY}{7.5}{\leftword}
  \ppowtext{2701}{\ppowY}{7.5}{\rightword}
}
\ppowtick{2598}{76}
\ppowtick{2598}{124}
\ppowcross{2598}{172}
\ppowcross{2598}{220}
\ppowcross{2598}{268}
\foreach \row/\leftword/\rightword in {
  0/Life/Life,1/is/is,2/a/roll,3/box/of,4/of/film}{
  \pgfmathsetmacro{\ppowY}{663+48*\row}
  \ppowtext{2500}{\ppowY}{7.5}{\leftword}
  \ppowtext{2692}{\ppowY}{7.5}{\rightword}
  \ppowtick{2598}{\ppowY+2}
}

\draw[ppowArrow] (2819,467) -- (2867,467);
\fill[ppowColumn,rounded corners=12bp] (2876,275) rectangle (3019,660);
\fill[ppowColumn,rounded corners=12bp] (3078,275) rectangle (3221,660);
\foreach \x in {2890,3100}{
  \draw[ppowSmall] (\x,297) rectangle (\x+103,367);
  \draw[ppowSmall] (\x,388) rectangle (\x+103,458);
  \draw[ppowSmall] (\x,479) rectangle (\x+103,549);
  \draw[ppowSmall] (\x,570) rectangle (\x+103,640);
}
\ppowfit{2941}{330.98}{14}{.800}{.951}{$r_1$}
\ppowfit{2943.42}{422.46}{14}{.961}{.927}{$r_2$}
\foreach \ppowDotY in {497,515,533}
  \fill[black] (2941,\ppowDotY) ellipse[x radius=1.5,y radius=2.5];
\ppowfit{2942.27}{601.50}{14}{.848}{.833}{$r_G$}
\ppowfit{3150.38}{332.88}{12}{.750}{.922}{$A_1$}
\ppowfit{3150.26}{421.76}{12}{.839}{.843}{$A_2$}
\foreach \ppowDotY in {498,516,534.5}
  \fill[black] (3149.5,\ppowDotY) ellipse[x radius=1.5,y radius=2.5];
\ppowfit{3150.29}{602.71}{12}{.787}{.804}{$A_G$}
\draw[ppowArrow] (3025,467) -- (3071,467);
\end{tikzpicture}%
}
\endgroup
\caption{\textbf{Overview of PPOW.} PPOW performs policy optimization at the window level for speculative decoding. \textbf{Left:} Adaptive windowing uses confidence-weighted draft--target divergence scores to prioritize informative training windows. \textbf{Right:} The drafter samples a candidate group of speculative windows for policy optimization with performance-driven Rewards and KL regularization.}
\label{fig:ppow_frame}
\end{figure*}

\section{Related Work}

Recent work has improved speculative decoding through advances in drafter design, draft--target alignment, and inference-time decoding optimization. One approach is to improve the drafter itself. Head-based methods, including MEDUSA~\cite{cai2024medusa} and Hydra~\cite{ankner2024hydra}, augment the model with auxiliary draft heads for multi-token prediction. Feature-based methods including EAGLE~\cite{li2024eagle, li2024eagle2, li2025eagle3}, HASS~\cite{zhang2025learning}, and GRIFFIN~\cite{hu2025griffin}, improve  drafting by leveraging hidden-state representations from the target model. Draft--target alignment has also been improved through distillation or online adaptation, as in DistillSpec~\cite{zhou2024distillspec},  OSD~\cite{liu2024online}, and FastDraft~\cite{zafrir2025fastdraft}. Lookahead decoding~\cite{fu2024break} provides a training-free alternative by constructing future candidates through Jacobi-style parallel decoding. These approaches are primarily trained with token-level prediction, distillation, or local distribution-matching objectives, rather than objectives defined over speculative windows.

Prior work has also improved speculative decoding at inference time through multi-candidate selection and verification algorithms~\cite{sun2023spectr, sun2024block}, tree-based speculative inference and verification~\cite{miao2024specinfer}, tree-structured and hardware-aware speculation~\cite{chen2024sequoia}, cascaded or adaptive speculation strategies~\cite{chen2024cascade, mamou2024dynamic, liu2025adaptive}, and parallel draft--verify execution mechanisms~\cite{svirschevski2024specexec, liu2024pearl}. These approaches mainly operate on the inference procedure, such as how draft candidates are constructed, organized, or verified, and are complementary to drafter-training methods.

Reward signals and reinforcement learning have also been explored in speculative decoding. RSD~\cite{liao2025reward} uses an inference-time reward model to guide draft acceptance, introducing a reward-guided verification mechanism that differs from standard rejection-sampling-based speculative decoding. Related efforts include SPEC-RL~\cite{liu2026specrl}, RLHFSpec~\cite{wang2025rlhfspec}, and ReSpec~\cite{chen2025respec}, which adapt speculative decoding to improve the efficiency of RL training rather than optimizing the drafter for standard speculative decoding inference.

The closest methods explicitly incorporate speculative acceptance into drafter training. LK Losses~\cite{samarin2026lk} design per-position losses to improve acceptance probabilities, while VSD~\cite{zou2026variational} formulates drafter training as variational inference over latent paths and updates the drafter with an acceptance-aware posterior in an EM-style procedure. In contrast, PPOW formulates drafter optimization as reinforcement learning over speculative windows, using performance-driven rewards derived from realized verification outcomes together with adaptive prioritization of hard-to-draft windows.

\section{Preliminaries}

We define the notation for speculative decoding used in this paper. Let $\pi_{\text{target}}$ and $\pi_{\theta}$ denote the target model and the draft model (drafter), respectively, where $\theta$ parameterizes the drafter. Given an input prefix $\boldsymbol{x}=(x_1,\dots,x_{n_1})$, the drafter proposes a speculative window of $K$ candidate tokens,
\[
\hat{\boldsymbol{y}}=(\hat{y}_1,\dots,\hat{y}_K).
\]
The target model then performs parallel verification with rejection sampling under the standard speculative decoding procedure, resulting in an accepted prefix of length $k \in \{0,\dots,K\}$.

\subsection{Feature-Based Drafting}

In this paper, we use feature-based drafters from the EAGLE family~\cite{li2024eagle, li2024eagle2, li2025eagle3} as the base drafters in our experiments. Feature-based drafters augment token-based drafting with hidden-state features from the target model. Specifically, at drafting step $t$, the drafter predicts the next-token distribution as
\[
\pi_{\theta}(\hat{y}_t \mid \boldsymbol{x}, \hat{\boldsymbol{y}}_{<t}, \mathbf{H}),
\]
where
\[
\mathbf{H} = \left[\mathbf{H}_{1:n_1}^{\text{tgt}}, \mathbf{H}_{<t}^{d}\right]
\]
denotes the concatenation of the target model's prefix hidden states $\mathbf{H}_{1:n_1}^{\text{tgt}}$ and the drafter's hidden states $\mathbf{H}_{<t}^{d}$ within the current speculative window. Further details of the rejection-sampling verification procedure are provided in Appendix~\ref{app:reject_sampling}.

\subsection{Policy Optimization over Speculative Windows}

In contrast to standard policy optimization formulations that operate on full responses, we formulate drafter training over fixed-length speculative windows. For each input prefix $\boldsymbol{x}$, the drafter defines a policy over speculative windows $\hat{\boldsymbol{y}}=(\hat{y}_1,\dots,\hat{y}_K)$, and a scalar reward $R(\boldsymbol{x}, \hat{\boldsymbol{y}})$ evaluates the window as a whole. The objective is
\[
J(\theta) = \mathbb{E}_{\hat{\boldsymbol{y}} \sim \pi_\theta}\!\left[R(\boldsymbol{x}, \hat{\boldsymbol{y}})\right].
\]

This formulation matches the speculative decoding procedure, in which a speculative window is verified as a whole and its utility is determined by the resulting accepted prefix. Appendix~\ref{app:sft_vs_rl} provides further discussion of window-level policy optimization versus token-level training for speculative decoding.

\section{PPOW}

PPOW is a performance-driven training framework for speculative drafters that optimizes window-level behavior for speculative decoding. Concretely, PPOW uses Adaptive Divergence-Aware Windowing to prioritize informative speculative windows, assigns rewards using the Cost-Aware Speedup Reward, complemented by the Distribution-Based Proximity Reward when accepted-prefix-based signals are sparse, and updates the drafter through window-level reinforcement learning. Figure~\ref{fig:ppow_frame} gives an overview of PPOW, and Algorithm~\ref{alg:ppow_algo} summarizes the full training procedure.

\subsection{Window-Level Reinforcement Learning for Speculative Decoding}
\label{sec:rl_framework}

PPOW formulates drafter training as reinforcement learning over speculative windows, so as to better align optimization with the inference mechanism of speculative decoding. In speculative decoding, the drafter produces speculative windows and the target model verifies them in parallel, yielding accepted-prefix outcomes for different speculative windows under the same prefix. Since these outcomes determine speculative efficiency, PPOW optimizes at the window level and learns from the relative speculative utility of speculative windows.

For a given input prefix $\boldsymbol{x}$, PPOW samples $G_{\mathrm{roll}}$ speculative windows $\hat{\boldsymbol{y}}=(\hat{y}_1,\dots,\hat{y}_K)$ from the drafter policy, forming a rollout group for the same prefix. This grouping can improve accepted-prefix outcomes by considering multiple speculative windows for the same prefix, and the extent of this improvement depends on the quality of the speculative windows in the group. PPOW therefore assigns each speculative window a scalar reward and normalizes rewards within the rollout group to produce group-relative advantages. Each speculative window is thus treated as a single training unit: it receives one window-level scalar reward, which is converted into a group-relative advantage. This grouped optimization provides a relative learning signal and remains aligned with practical speculative decoding settings that consider multiple speculative windows.

In addition, PPOW incorporates a KL regularization term into the objective. We treat the drafter as the policy $\pi_{\theta}$ and the target model $\pi_{\text{target}}$ as a cross-scale distributional anchor, and use the KL divergence $D_{\mathrm{KL}}(\pi_{\theta}\,\|\,\pi_{\text{target}})$ to encourage alignment during RL exploration. This target-anchored regularization is intended to stabilize training while keeping policy updates aligned with the target distribution used for speculative verification.

The resulting objective can be written as
\[
\begin{split}
J(\theta) = &\frac{1}{G_{\mathrm{roll}}} \sum_{i=1}^{G_{\mathrm{roll}}} \frac{1}{K} \sum_{t=1}^{K} \Big[
\min \Big(
r_{i,t}(\theta)\hat{A}_i,\,
\mathrm{clip}\big(r_{i,t}(\theta), 1-\epsilon_{\mathrm{clip}}, 1+\epsilon_{\mathrm{clip}}\big)\hat{A}_i
\Big) \\
&\qquad - \beta\, D_{\mathrm{KL}}\!\big(
\pi_{\theta}(\cdot \mid \boldsymbol{x}, \hat{\boldsymbol{y}}_{i,<t}, \mathbf{H})
\,\|\, 
\pi_{\mathrm{target}}(\cdot \mid \boldsymbol{x}, \hat{\boldsymbol{y}}_{i,<t})
\big)
\Big],
\end{split}
\]
where
\[
r_{i,t}(\theta)=
\frac{\pi_{\theta}(\hat{y}_{i,t}\mid \boldsymbol{x}, \hat{\boldsymbol{y}}_{i,<t}, \mathbf{H})}
{\pi_{\mathrm{old}}(\hat{y}_{i,t}\mid \boldsymbol{x}, \hat{\boldsymbol{y}}_{i,<t}, \mathbf{H})},
\]
and $\hat{A}_{i}$ denotes the normalized group-relative advantage of the $i$-th speculative window, capturing its relative speculative utility within the rollout group. Specifically, for a rollout group of speculative windows sampled under the same prefix, we compute
\[
\hat A_i = \frac{R_i - \mu_R}{\sigma_R + \delta},
\]
where \(R_i\) is the scalar reward of the \(i\)-th speculative window, and \(\mu_R\) and \(\sigma_R\) are the mean and standard deviation of rewards within the rollout group.

Although the policy is autoregressively factorized over tokens, optimization is performed at the speculative-window level, with all tokens within the same window sharing a common advantage signal.

\subsection{Adaptive Divergence-Aware Windowing}
\label{sec:adaptive_sampling}

Window-level training better matches speculative decoding at inference time, but it also leads to substantial sample redundancy. A full response can be decomposed into many overlapping speculative windows, and many of them are already adequately modeled by a supervised-initialized drafter. At the same time, speculative performance is often constrained by certain hard-to-draft positions, where draft--target mismatch can more strongly affect the accepted prefix. PPOW therefore prioritizes speculative windows that are both more informative for training and more consequential for speculative acceptance length through Adaptive Divergence-Aware Windowing (ADAW).

Let $P_t$ and $Q_t$ denote the target and drafter distributions at position $t$,
\[
P_t = \pi_{\text{target}}(\cdot \mid \boldsymbol{x}, \boldsymbol{y}_{<t}),
\qquad
Q_t = \pi_{\theta}(\cdot \mid \boldsymbol{x}, \boldsymbol{y}_{<t}, \mathbf{H}),
\]
and define the token-level \emph{criticality score} as
\[
v_t = C(P_t)\cdot D_{\mathrm{KL}}(P_t \,\|\, Q_t),
\]
where
\[
C(P_t)=1-\frac{H(P_t)}{\log |\mathcal{V}|},
\]
and $\mathcal{V}$ denotes the vocabulary, and $H(P_t)$ is the Shannon entropy of the target distribution $P_t$.

The criticality score $v_t$ is a confidence-weighted measure of draft--target divergence. Here, $D_{\mathrm{KL}}(P_t \,\|\, Q_t)$ captures mismatch between the drafter and target distributions, while $C(P_t)$ emphasizes contexts in which the target distribution is more concentrated and speculative verification is more sensitive to such mismatch. In this way, ADAW emphasizes contexts that are both difficult for the drafter and more consequential for speculative acceptance.

Because speculative decoding operates on complete windows, PPOW aggregates token-level criticality over each speculative window:
\[
s_j = \frac{1}{K}\sum_{t=j}^{j+K-1} v_t,
\]
and prioritizes windows with larger $s_j$ during training. This adaptive windowing strategy reduces redundancy while focusing optimization on windows more directly associated with acceptance bottlenecks. Appendix~\ref{app:proofs} analyzes the connection between draft--target divergence and speculative acceptance behavior, and further discusses the role of confidence weighting in prioritizing training windows.

\subsection{Performance-Driven Rewards}
\label{sec:performance_rewards}

PPOW uses a window-level reward with two components. The primary component is a Cost-Aware Speedup Reward that encourages longer accepted prefixes while accounting for both drafting and verification cost. The second component is a complementary Distribution-Based Proximity Reward that provides auxiliary credit when exact verification terminates early but the speculative window still achieves a similar cumulative target-model log-likelihood.

\subsubsection{Cost-Aware Speedup Reward}

The primary goal of speculative decoding is to improve inference efficiency. PPOW therefore uses a cost-aware reward based on the accepted prefix length $k$ of a speculative window and the relative cost $\gamma$ of the drafter:
\[
R_{\text{speedup}} = \frac{k}{k\gamma + 1},
\]
where $k\in\{0,\dots,K\}$ is the accepted length and $\gamma$ denotes the relative computational cost of $\pi_\theta$ with respect to $\pi_{\text{target}}$, estimated in our implementation by the ratio of non-embedding parameters.

This formulation accounts for the computational structure of speculative decoding by balancing accepted length against relative drafting cost. The term $k\gamma + 1$ combines the drafting cost and the target-model verification cost within a speculative step. Compared with using raw the accepted prefix length $k$ alone, $R_{\text{speedup}}$ provides a cost-aware objective intended to better reflect the efficiency trade-off in speculative decoding.

We use this formulation during training instead of rewards derived from directly measured speedup, since measured speedup depends on the execution environment and is therefore less suitable as a general speculative-window-level training reward. Appendix~\ref{app:speedup_proxy} compares the Cost-Aware Speedup Reward with a Measured-Speedup-Based alternative and shows that it preserves the same acceptance-related trend while remaining effective for optimization.

\subsubsection{Distribution-Based Proximity Reward}

The accepted-prefix reward alone can become sparse when verification terminates early, even if the drafted window remains broadly compatible with the target model's preferences. To provide auxiliary partial credit in such cases, PPOW introduces the Distribution-Based Proximity Reward, which compares the drafted window with a target-preferred window under the target model's cumulative log-likelihood. As shown in Figure~\ref{fig:token_mismatch}(b), early verification truncation yields $k=0$ when the drafter and target model produce different textual realizations of similar content.

For a drafted speculative window $\hat{\boldsymbol{y}}$, we construct a target-preferred reference window \(\boldsymbol{y}\) autoregressively from the target model, and compare their cumulative log-probabilities under the target model:
\[
\Delta =
\sum_{t=1}^{K}
\Big[
\log \pi_{\text{target}}(y_t \mid \boldsymbol{x}, \boldsymbol{y}_{<t})
-
\log \pi_{\text{target}}(\hat{y}_t \mid \boldsymbol{x}, \hat{\boldsymbol{y}}_{<t})
\Big],
\]
where
\[
y_t = \arg\max_y \pi_{\text{target}}(y \mid \boldsymbol{x}, \boldsymbol{y}_{<t}).
\]
We then define
\[
R_{\text{dist}} = \eta \cdot \mathbf{1}[\Delta<\epsilon],
\]
where $\epsilon$ is a tolerance threshold, and $\eta$ scales the contribution of this auxiliary reward. Appendix~\ref{app:eta_sensitivity} studies sensitivity to the choice of $\eta$.

This auxiliary reward is designed to capture training signal that may be missed by accepted-prefix-based reward alone. Rather than serving as a primary optimization target, \(R_{\text{dist}}\) provides bounded partial credit to drafted windows whose cumulative target log-probabilities remain close to those of the target-preferred reference window. This term supplies additional signal while preventing the auxiliary term from dominating the primary speedup objective.

As a result, $R_{\text{dist}}$ complements the accepted-prefix reward with a softer window-level signal and can provide denser feedback during training. Section~\ref{sec:ablation} evaluates its overall contribution in ablations, and Appendix~\ref{app:easy_hard_case} further examines its effect on easy and hard windows.

\section{Experiments}

We evaluate PPOW on end-to-end speculative decoding performance across multiple model families and diverse task settings, compare it with continued supervised training, examine its behavior under different inference candidate group sizes, and ablate its reward design and divergence-aware windowing strategy. Additional analyses of the speedup proxy, easy/hard-window behavior, and broader baseline comparisons are provided in the appendix.

\subsection{Experimental Setup}
\label{sec:experimental_setup}

\paragraph{Models and Tasks.}
We evaluate PPOW on two model families: LLaMA-3~\cite{grattafiori2024llama} (8B, 70B) and Qwen3~\cite{yang2025qwen3} (8B, 32B). PPOW is applied to feature-based EAGLE family~\cite{li2024eagle, li2024eagle2, li2025eagle3}, initialized from supervised checkpoints before RL optimization. We evaluate on MT-Bench~\cite{zheng2023judging} for multi-turn dialogue, HumanEval~\cite{chen2021evaluating} for code generation, and GSM8K~\cite{cobbe2021training} for mathematical reasoning.

\paragraph{Metrics.}
We report two standard metrics for speculative decoding:
\begin{itemize}[leftmargin=3em, nosep, parsep=1pt]
    \item \textbf{Average Acceptance Length ($\tau$):} the average number of drafted tokens accepted per speculative verification step.
    \item \textbf{Speedup Ratio:} end-to-end wall-clock speedup relative to standard autoregressive decoding under the same hardware and software setup.
\end{itemize}

\paragraph{Baselines.}
We mainly compare with learned drafters, including EAGLE-3~\cite{li2025eagle3} and GRIFFIN~\cite{hu2025griffin}. Optimization details and the unified decoding protocol are provided in Appendix~\ref{app:optimization_implementation_details}, and additional results for OSD~\cite{liu2024online} and FastDraft~\cite{zafrir2025fastdraft} are provided in Appendix~\ref{app:more_baseline}.

\subsection{Main Results}
\label{sec:main_results}

Table~\ref{tab:sota_comparison} presents our main end-to-end speculative decoding results under a unified decoding protocol. The table covers two model families (LLaMA-3 and Qwen3), multiple model scales (8B, 32B, and 70B), three representative benchmarks spanning dialogue, code generation, and mathematical reasoning, and two temperature settings, while reporting both average acceptance length and end-to-end wall-clock speedup. Across this broad evaluation setup, PPOW achieves the best mean results for every model family under both temperature settings.

\begin{table}[htbp]
\centering
\caption{\textbf{Main speculative decoding results.} Average acceptance length ($\tau$) and speedup across models and benchmarks under a unified decoding protocol. L31, L33, and Q3 refer to LLaMA-3.1-Instruct, LLaMA-3.3-Instruct, and Qwen3, respectively.}
\label{tab:sota_comparison}
\small
\setlength{\tabcolsep}{3pt}
\renewcommand{\arraystretch}{1.0}
\begin{tabular*}{0.95\textwidth}{@{\extracolsep{\fill}}ll cc cc cc cc}
\toprule
\multirow{2}{*}{Model} & \multirow{2}{*}{Method} & \multicolumn{2}{c}{MT-Bench} & \multicolumn{2}{c}{HumanEval} & \multicolumn{2}{c}{GSM8K} & \multicolumn{2}{c}{Mean} \\
\cmidrule(lr){3-4} \cmidrule(lr){5-6} \cmidrule(lr){7-8} \cmidrule(lr){9-10}
& & $\tau$ & Speedup & $\tau$ & Speedup & $\tau$ & Speedup & $\tau$ & Speedup \\
\midrule
\multicolumn{10}{c}{Temperature = 0.0} \\
\midrule
\multirow{3}{*}{L31-8B}
 & GRIFFIN & 5.14 & 2.58$\times$ & 6.72 & 3.96$\times$ & 5.98 & 3.38$\times$ & 5.95 & 3.31$\times$ \\[0.2pt]
& EAGLE-3 & \textbf{5.53} & \textbf{2.91$\times$} & 6.63 & 3.92$\times$ & 6.12 & 3.41$\times$ & 6.09 & 3.41$\times$ \\[0.2pt]
& PPOW & 5.47 & 2.72$\times$ & \textbf{7.23} & \textbf{4.14$\times$} & \textbf{6.50} & \textbf{3.52$\times$} & \textbf{6.40} & \textbf{3.46$\times$} \\
\cmidrule{1-10}
\multirow{3}{*}{L33-70B}
& GRIFFIN & 5.08 & 3.60$\times$ & 6.69 & 4.78$\times$ & 5.90 & 3.99$\times$ & 5.89 & 4.12$\times$ \\[0.2pt]
& EAGLE-3 & 5.12 & 3.63$\times$ & 6.78 & 4.80$\times$ & 5.93 & 4.02$\times$ & 5.94 & 4.15$\times$ \\[0.2pt]
& PPOW & \textbf{5.45} & \textbf{3.73$\times$} & \textbf{6.96} & \textbf{4.82$\times$} & \textbf{6.47} & \textbf{4.54$\times$} & \textbf{6.29} & \textbf{4.36$\times$} \\
\cmidrule{1-10}
\multirow{2}{*}{Q3-8B}
& EAGLE-3 & 4.95 & 2.64$\times$ & 6.68 & 3.40$\times$ & 6.86 & 3.47$\times$ & 6.16 & 3.17$\times$ \\[0.2pt]
& PPOW & \textbf{5.58} & \textbf{3.02$\times$} & \textbf{7.01} & \textbf{3.62$\times$} & \textbf{6.97} & \textbf{3.54$\times$} & \textbf{6.52} & \textbf{3.39$\times$} \\
\cmidrule{1-10}
\multirow{2}{*}{Q3-32B}
& EAGLE-3 & 5.25 & 3.47$\times$ & 6.52 & 4.02$\times$ & 6.21 & 3.94$\times$ & 5.99 & 3.81$\times$ \\[0.2pt]
& PPOW & \textbf{5.78} & \textbf{3.54$\times$} & \textbf{6.91} & \textbf{4.16$\times$} & \textbf{6.62} & \textbf{4.05$\times$} & \textbf{6.44} & \textbf{3.92$\times$} \\
\midrule
\multicolumn{10}{c}{Temperature = 1.0} \\
\midrule
\multirow{3}{*}{L31-8B}
& GRIFFIN & 4.03 & 2.24$\times$ & 6.01 & 3.29$\times$ & 5.13 & 2.89$\times$ & 5.06 & 2.81$\times$ \\[0.2pt]
& EAGLE-3 & \textbf{4.17} & \textbf{2.35$\times$} & 5.94 & 3.25$\times$ & 5.16 & 2.89$\times$ & 5.09 & 2.83$\times$ \\[0.2pt]
& PPOW & 4.12 & 2.29$\times$ & \textbf{6.33} & \textbf{3.37$\times$} & \textbf{5.63} & \textbf{3.13$\times$} & \textbf{5.36} & \textbf{2.93$\times$} \\
\cmidrule{1-10}
\multirow{3}{*}{L33-70B}
& GRIFFIN & 4.70 & 3.38$\times$ & 6.31 & 4.46$\times$ & 5.54 & 3.87$\times$ & 5.52 & 3.90$\times$ \\[0.2pt]
& EAGLE-3 & 4.82 & 3.40$\times$ & 6.28 & 4.43$\times$ & 5.98 & 4.16$\times$ & 5.69 & 4.00$\times$ \\[0.2pt]
& PPOW & \textbf{5.14} & \textbf{3.52$\times$} & \textbf{6.57} & \textbf{4.62$\times$} & \textbf{6.29} & \textbf{4.49$\times$} & \textbf{6.00} & \textbf{4.21$\times$} \\
\cmidrule{1-10}
\multirow{2}{*}{Q3-8B}
& EAGLE-3 & 4.00 & 2.41$\times$ & 5.71 & 2.11$\times$ & 5.52 & 2.89$\times$ & 5.08 & 2.47$\times$ \\[0.2pt]
& PPOW & \textbf{4.72} & \textbf{2.82$\times$} & \textbf{5.96} & \textbf{2.33$\times$} & \textbf{5.76} & \textbf{3.01$\times$} & \textbf{5.48} & \textbf{2.72$\times$} \\
\cmidrule{1-10}
\multirow{2}{*}{Q3-32B}
& EAGLE-3 & 4.51 & 2.60$\times$ & 5.53 & 2.59$\times$ & 5.23 & 2.81$\times$ & 5.09 & 2.67$\times$ \\[0.2pt]
& PPOW & \textbf{4.95} & \textbf{2.81$\times$} & \textbf{6.17} & \textbf{2.77$\times$} & \textbf{5.61} & \textbf{3.01$\times$} & \textbf{5.58} & \textbf{2.86$\times$} \\
\bottomrule
\end{tabular*}
\end{table}

At temperature $0.0$, PPOW achieves mean acceptance lengths of 6.29--6.52 and mean speedups of 3.39--4.36$\times$ across the evaluated model families, with consistent gains also observed under temperature $1.0$. The consistent gains at temperature $1.0$ further indicate that PPOW remains beneficial beyond near-greedy decoding, even under less concentrated output distributions. Beyond the mean results, PPOW improves both acceptance length and wall-clock speedup on most benchmark--model combinations, showing that PPOW improves drafting behavior in a way that translates into practical end-to-end efficiency gains rather than acceptance-only improvements.

The gains are especially pronounced on HumanEval and GSM8K. For example, on LLaMA-3.1-8B, PPOW improves HumanEval from $\tau=6.63$ and $3.92\times$ speedup relative to EAGLE-3 to $\tau=7.23$ and $4.14\times$, while on GSM8K it improves from $\tau=6.12$ and $3.41\times$ to $\tau=6.50$ and $3.52\times$. On MT-Bench, the improvements are more modest, which may be related to the higher diversity and uncertainty of valid continuations in open-ended dialogue. Additional results on broader natural-language tasks and supplementary comparisons with OSD~\cite{liu2024online} and FastDraft~\cite{zafrir2025fastdraft} are provided in Appendix~\ref{app:more_baseline}.

\subsection{PPOW versus Continued Supervised Training}
\label{sec:sft_comparison}

We compare PPOW with continued supervised training (CST) from the same EAGLE-3~\cite{li2025eagle3} initialization. This allows us to examine both whether PPOW's gains can be matched by additional post-training compute under continued token-level training, and whether any remaining difference is better explained by the training objective itself. For CST, we consider two learning rates: 5e-5 from the EAGLE-3 setup and a smaller 5e-6 as a conservative control to probe whether continued token-level training can still yield further gains. For PPOW, we report the best-performing learning rate. Both methods start from the same checkpoint and are trained for matched numbers of update steps, the comparison helps distinguish gains from additional training alone from gains due to optimization that is more directly aligned with speculative acceptance behavior.

\begin{figure}[htbp]
  \centering
  \includegraphics[width=0.80\textwidth]{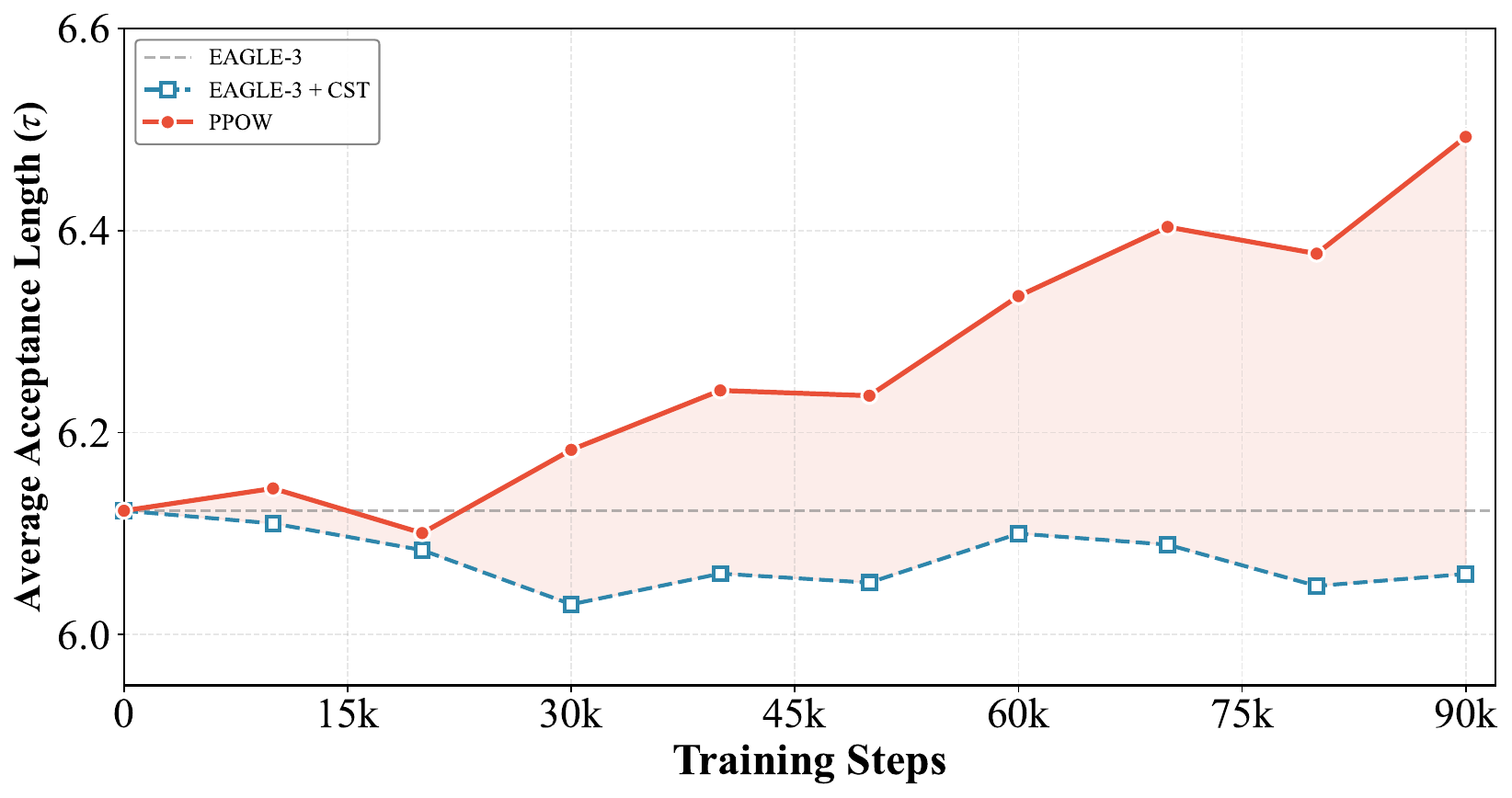}
  \caption{\textbf{PPOW versus CST under matched training steps.} On GSM8K with LLaMA-3.1-8B, CST with learning rate 5e-6 fluctuates around the original checkpoint's average acceptance length, while PPOW steadily improves throughout training.}
  \label{fig:training_efficiency}
\end{figure}

\begin{table}[htbp]
\centering
\caption{\textbf{Average acceptance length ($\tau$) under continued training on GSM8K with LLaMA-3.1-8B.} Continued supervised training yields limited gains across both learning rates, while PPOW improves more steadily.}
\label{tab:performance_gains}
\vspace{-2.1pt}
\small
\begin{tabular*}{0.80\columnwidth}
{@{\hspace{5pt}\extracolsep{\fill}}cccc@{\hspace{5pt}}}
\toprule
\multirow{2}{*}{steps} & \multicolumn{2}{c}{EAGLE-3 (CST)} & \multicolumn{1}{c}{PPOW} \\
\cmidrule(lr){2-3} \cmidrule(lr){4-4}
& $5\times10^{-5}$ & $5\times10^{-6}$ & $5\times10^{-6}$ \\
\midrule
30k & 5.47 & 6.03 & 6.12 \\[1.2pt]
60k & 5.61 & 6.10 & 6.32 \\[1.2pt]
90k & 5.52 & 6.06 & \textbf{6.49} \\[1.2pt]
\bottomrule
\end{tabular*}
\end{table}

Table~\ref{tab:performance_gains} reports GSM8K results with LLaMA-3.1-8B for CST at both learning rates and for PPOW under its best-performing learning rate, all starting from the same EAGLE-3 checkpoint. Over 30k / 60k / 90k training steps, CST with 5e-6 fluctuates around the original checkpoint's average acceptance length, reaching 6.03 / 6.10 / 6.06, while PPOW improves more steadily to 6.12 / 6.32 / 6.49. The higher CST learning rate of 5e-5 shows the same overall pattern. Figure~\ref{fig:training_efficiency} further illustrates this difference in training behavior, with CST remaining nearly flat while PPOW continues to improve. 

These results indicate that EAGLE-3 is already close to saturation under continued token-level supervised training, with only limited gains from further optimization under the same objective. In contrast, PPOW benefits from optimizing an objective that is more directly aligned with acceptance length in speculative decoding. We discuss the objective-level difference between PPOW and continued supervised training in Appendix~\ref{app:sft_vs_rl}, and observe the same qualitative trend on natural-language tasks, as shown in Appendix~\ref{app:nl_tasks}.

\subsection{Trade-offs with Different Inference Candidate Group Sizes}
\label{sec:throughput_groupsize}

Table~\ref{tab:groupsize_tradeoff} evaluates the same trained PPOW model and the EAGLE-3~\cite{li2025eagle3} across different inference candidate group sizes. In practice, speculative decoding often generates multiple candidate sequences via branching or tree decoding and verifies them in parallel. Larger candidate groups generally increase the chance of obtaining a longer accepted prefix, but they also incur higher verification cost. PPOW reaches strong acceptance with smaller candidate groups. At candidate group size 4, PPOW achieves $\tau=6.33$, while EAGLE-3 reaches only $\tau=5.03$ under the same setting and requires a group size of 16 to approach PPOW's performance. This result shows that PPOW can achieve better acceptance under a much smaller verification budget, highlighting its practical advantage in inference settings where verification overhead matters.

\begin{table}[htbp]
\centering
\caption{\textbf{Average acceptance length ($\tau$) under different inference candidate group sizes.} PPOW achieves higher acceptance with smaller candidate groups than the supervised baseline on LLaMA-3.1-8B / GSM8K.}
\label{tab:groupsize_tradeoff}
\small
\begin{tabular*}{0.80\columnwidth}
{@{\hspace{5pt}\extracolsep{\fill}}ccc@{\hspace{5pt}}}
\toprule
Candidate Group Size & PPOW & EAGLE-3 \\
\midrule
1  & 4.92 & 4.10 \\[0.5pt]
2   & 5.23 & 4.46 \\[0.5pt]
4 & \textbf{6.33} & 5.03 \\[0.5pt]
8 & 6.41 & 5.58 \\[0.5pt]
16 & 6.49 & \textbf{6.12} \\
\bottomrule
\end{tabular*}
\end{table}

\subsection{Ablation Studies}
\label{sec:ablation}

We examine the contribution of PPOW's key components beyond the base speedup reward: the Distribution-Based Proximity Reward, Adaptive Divergence-Aware Windowing, and the confidence factor $C(P_t)$ used in ADAW. Table~\ref{tab:ablation} reports component ablations on LLaMA-3.1-8B.

\begin{table}[h]
\centering
\caption{\textbf{Ablation of PPOW components on LLaMA-3.1-8B.} ``w/o $R_{\text{dist}}$'' removes the Distribution-Based Proximity Reward, ``w/o $C(P_t)$'' removes the confidence factor in ADAW and uses divergence-only window prioritization, ``w/o ADAW'' replaces ADAW with uniform window sampling, and ``w/o both'' removes $R_{\text{dist}}$ and uses uniform window sampling.}
\label{tab:ablation}
\small
\setlength{\tabcolsep}{2pt}
\renewcommand{\arraystretch}{1.20}
\begin{tabular*}{0.80\columnwidth}{@{\hspace{5pt}\extracolsep{\fill}}lcccc@{\hspace{5pt}}}
\toprule
\multirow{2}{*}{\normalsize Method} & \multicolumn{2}{c}{MT-Bench} & \multicolumn{2}{c}{GSM8K} \\
\cmidrule(lr){2-3} \cmidrule(lr){4-5}
& $\tau$ & Speedup & $\tau$ & Speedup \\
\midrule
w/o $R_{\text{dist}}$ & 5.05 & 2.52$\times$ & 6.41 & 3.48$\times$ \\
w/o $C(P_t)$ & 4.47 & 2.39$\times$ & 5.92 & 3.12$\times$ \\
w/o ADAW & 4.82 & 2.44$\times$ & 6.35 & 3.37$\times$ \\
w/o both & 4.38 & 2.39$\times$ & 6.05 & 3.26$\times$ \\
\midrule
\textbf{PPOW (Full)} & \textbf{5.47} & \textbf{2.72$\times$} & \textbf{6.50} & \textbf{3.52$\times$} \\
\bottomrule
\end{tabular*}
\end{table}

Removing either $R_{\text{dist}}$ or ADAW reduces both acceptance length and speedup, and removing both further degrades performance on both benchmarks. On MT-Bench, removing both reduces $\tau$ from 5.47 to 4.38 and speedup from 2.72$\times$ to 2.39$\times$. This suggests that the two components are complementary in practice: $R_{\text{dist}}$ provides a denser learning signal when exact verification is sparse, and ADAW improves training efficiency by focusing optimization on more informative windows.

We further evaluate a divergence-only variant of ADAW by removing the confidence factor $C(P_t)$. This variant performs worse than both full ADAW and uniform window sampling. On GSM8K, removing $C(P_t)$ lowers $\tau$ from 6.50 to 5.92, compared with 6.35 under uniform sampling. This highlights the role of confidence weighting in encouraging ADAW to focus on bottleneck windows rather than ambiguous high-entropy mismatches. 

\begin{figure}[htbp]
  \centering
  \includegraphics[width=0.80\textwidth]{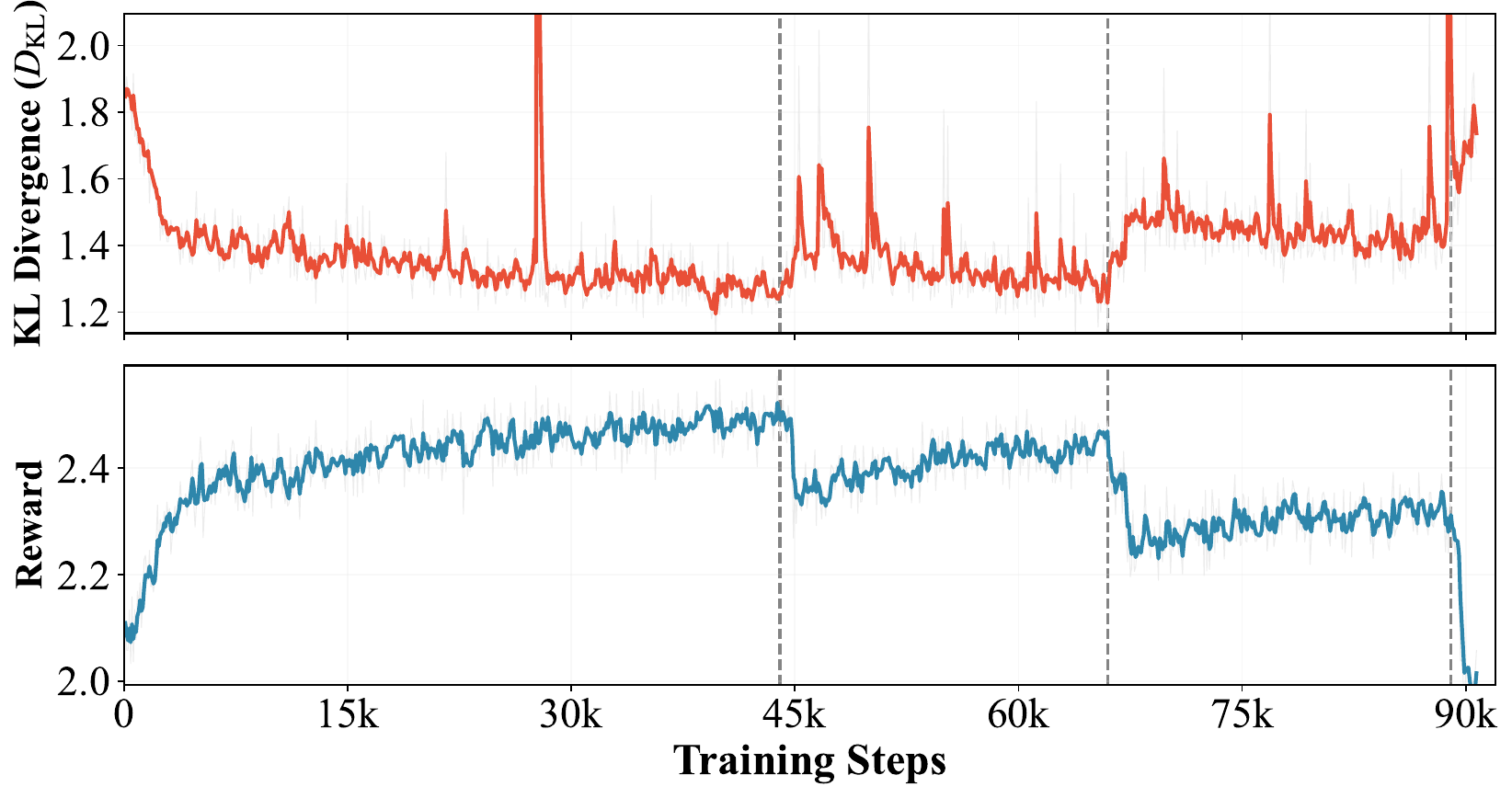}
  \caption{\textbf{Training dynamics under ADAW.} The dashed lines mark stage transitions where the hard-window ratio increases from 0.2 to 0.4 and then to 0.6. As more hard-to-draft windows are sampled, reward decreases and KL divergence rises.}
  \label{fig:kl_reward_vertical}
\end{figure}

Figure~\ref{fig:kl_reward_vertical} further illustrates the training dynamics under ADAW. The dashed lines mark stage transitions in the hard-window curriculum. After 44k steps, training switches from uniform window sampling to ADAW-based sampling, and in later stages the hard-window ratio is further increased from 0.2 to 0.4 and then to 0.6, making the sampled training windows progressively harder. As a result, average reward exhibits stage-wise drops while KL divergence rises, reflecting shifts in the sampled training window distribution toward hard-to-draft windows rather than optimization failure. Within each stage, however, the policy gradually adapts to the harder sampled windows, and performance continues to improve as training proceeds. 

Appendix~\ref{app:adaw_curriculum} describes the hard-window curriculum and the role of confidence weighting, and Appendix~\ref{app:proofs} provides analytical support for weighting draft--target divergence by target confidence. Appendix~\ref{app:easy_hard_case} provides additional ablations and easy/hard-window analysis of the Distribution-Based Proximity Reward, and Appendix~\ref{app:eta_sensitivity} studies sensitivity to its weight $\eta$. Appendix~\ref{app:limitations and impacts} discusses the limitations and broader impacts of PPOW.

\section{Conclusion}
We introduced PPOW, a performance-driven reinforcement learning framework for speculative decoding that optimizes the drafter over speculative windows rather than individual tokens. PPOW combines performance-driven rewards and adaptive hard-to-draft window selection to better align drafter optimization with inference-time acceptance behavior. Across diverse settings, PPOW consistently improves both average acceptance length and speedup. It also remains effective with substantially smaller candidate group sizes at inference time, highlighting its practical value under constrained verification budgets. In addition, we provide analytical support for the confidence-weighted draft--target divergence criterion used in ADAW. Overall, our results show that performance-driven, window-level drafter optimization is an effective way to improve speculative decoding efficiency.

\bibliographystyle{unsrt} 
\bibliography{arxiv_2026}

\newpage
\appendix

\section{Optimization and Implementation Details}
\label{app:optimization_implementation_details}

This appendix summarizes the training setup, optimization procedure, hyperparameter settings, and full PPOW training algorithm used in our experiments.

\subsection{Training Setup}
\label{app:training_setup}
Our training uses a two-stage pipeline: supervised initialization followed by PPOW-based reinforcement learning. For LLaMA-3, we initialize from the official EAGLE-3 checkpoints~\cite{li2025eagle3}. For Qwen, we first train the feature-based drafter with supervised fine-tuning on ShareGPT~\cite{vicuna2023} and UltraChat-200k~\cite{ding2023enhancing}, then further optimize it with PPOW on the same data mixture. All experiments are run on NVIDIA H100 (80GB) GPUs using PyTorch with FSDP~\cite{zhao2023pytorch} to support a frozen target model and a trainable drafter. PPOW training costs about 50, 100, and 200 GPU-hours for 8B, 32B, and 70B targets, respectively. 

\subsection{Policy Optimization and Target-Anchored KL Regularization}
\label{app:ppow_kl_regular}
PPOW is implemented with a group-relative clipped policy objective~\cite{schulman2017proximal, shao2024deepseekmath}. For each selected speculative window, we sample a rollout group of $G_{\mathrm{roll}}$ speculative windows, compute a scalar reward for each draft, and normalize rewards within the rollout group to obtain group-relative advantages. A single normalized advantage is then applied across the token-level updates within that speculative window. Unless otherwise specified, we do not use a separate value function.

A key difference from standard RL training setups lies in the KL regularization. Rather than constraining the current policy toward an initialization policy or an auxiliary reference model, PPOW computes KL regularization between the drafter policy and the frozen target model. Specifically, we compute the token-wise KL divergence between the draft and target distributions at each drafted position and average it over the speculative window. This keeps exploration anchored to the distribution that ultimately governs speculative verification.

\subsection{Training and Decoding Configurations}
Table~\ref{tab:hyperparams} summarizes the default training hyperparameters, hard-window curriculum schedule, and decoding settings used throughout our experiments. Under hard-window curriculum scheduling, PPOW begins with a smaller proportion of ADAW-selected hard windows and gradually increases this proportion over the course of training.

\begin{table}[htbp]
\centering
\caption{Training hyperparameters, curriculum schedule, and decoding configurations used in our experiments.}
\label{tab:hyperparams}
\begin{tabular*}{0.80\columnwidth}{@{\extracolsep{\fill}}ll@{\extracolsep{\fill}}}
\toprule
\textbf{Hyperparameters} & \textbf{Value} \\
\midrule
\rowcolor[gray]{0.95} \multicolumn{2}{l}{\textit{Training Configuration}} \\
\quad Clip ratio $\epsilon_{\text{clip}}$ & 0.2 \\
\quad KL coefficient $\beta$ & 0.03 \\
\quad Rollout group size $G_{\mathrm{roll}}$ & 8 \\
\quad Rollout / Sampling temperature & 1.0 \\
\quad Window size $K$ & 10 \\
\quad Hard-window curriculum & 0.2 / 0.4 / 0.6 \\
\quad Learning rate & $5\times 10^{-6}$ \\
\quad Warmup ratio & 0.05 \\
\quad Precision & bf16 \\
\midrule
\rowcolor[gray]{0.95} \multicolumn{2}{l}{\textit{Decoding Configurations}} \\
\quad Decoding strategy & tree decoding \\
\quad KV cache & true \\
\quad Speculative depth / num\_steps & 10 \\
\quad Branching factor / top-k & 3 \\
\quad Verification rule & rejection sampling \\
\quad Temperatures & 0.0 / 1.0 \\
\bottomrule
\end{tabular*}
\end{table}

\subsection{Hard-Window Curriculum in ADAW}
\label{app:adaw_curriculum}
ADAW's curriculum designed to avoid overwhelming the policy too early. For the hard-window curriculum, the $0.2 \rightarrow 0.4 \rightarrow 0.6$ schedule is used to gradually increase the fraction of hard windows during training. If the proportion of hard windows is increased too aggressively early in training, accepted prefixes tend to be very short and many rollout groups provide little useful relative signal, which slows training progress. Gradually increasing the hard-window ratio allows PPOW to first adapt on easier windows and then increasingly focus on the true bottlenecks. This is also consistent with the training dynamics already shown in Figure~\ref{fig:kl_reward_vertical}, where enabling ADAW initially lowers reward and raises KL before the policy adapts to the harder window distribution.

The confidence factor $C(P_t)$ helps distinguish \textbf{true bottlenecks} from \textbf{ambiguous high-entropy mismatches}. When the target distribution is concentrated, large draft--target divergence is more likely to indicate a consequential bottleneck, since an early mismatch is more likely to truncate the acceptance chain. Prioritizing such positions encourages the drafter to correct errors with larger impact on speculative efficiency.

When the target distribution is diffuse, however, divergence alone can over-prioritize positions where the target itself has no strong preference. Confidence weighting suppresses these noisy cases and keeps ADAW focused on windows that matter more for speculative efficiency.

\subsection{Full PPOW Training Algorithm}
\begin{algorithm}[htbp]
\caption{PPOW Training Procedure}
\label{alg:ppow_algo}
\begin{algorithmic}[1]
\STATE \textbf{Inputs:} drafter $\pi_{\theta}$, target model $\pi_{\text{target}}$, window size $K$, group size $G_{\mathrm{roll}}$, clip ratio $\epsilon_{\mathrm{clip}}$, KL coefficient $\beta$, relative cost $\gamma$, proximity threshold $\epsilon$, proximity weight $\eta$
\FOR{each training prefix $\boldsymbol{x}$}
\STATE {\color{green!50!black}$\triangleright$ \textit{Generate target-side outputs and draft distributions}}
    \STATE \hspace{1em} $(\boldsymbol{y}, \mathbf{H}, \{P_t\}_{t=1}^{T}) \leftarrow \pi_{\text{target}}(\boldsymbol{x})$
    \STATE \hspace{1em} $\{Q_t\}_{t=1}^{T} \leftarrow \pi_{\theta}(\boldsymbol{x}, \boldsymbol{y}, \mathbf{H})$

\STATE {\color{green!50!black}$\triangleright$ \textit{Compute criticality score}}
    \FOR{$t=1,\dots,T$}
        \STATE $v_t \leftarrow \Big(1-\frac{H(P_t)}{\log |\mathcal{V}|}\Big)\cdot D_{\mathrm{KL}}(P_t \,\|\, Q_t)$
    \ENDFOR
\STATE {\color{green!50!black}$\triangleright$ \textit{ADAW-based window selection}}
    \FOR{each window start index $j \in \{1,\dots,T-K+1\}$}
        \STATE $s_j \leftarrow \frac{1}{K}\sum_{t=j}^{j+K-1} v_t$
    \ENDFOR
    \STATE Sample a training window start $j^\star$ from the normalized weights proportional to $\{s_j\}$

\STATE {\color{green!50!black}$\triangleright$ \textit{Grouped rollout and reward computation}}
\FOR{$g=1,\dots,G_{\mathrm{roll}}$}
    \STATE Rollout a speculative window $\hat{\boldsymbol{y}}_g \sim \pi_{\theta}(\cdot \mid \boldsymbol{x}, \boldsymbol{y}_{<j^\star}, \mathbf{H}_{<j^\star})$
    \STATE $k_g \leftarrow \textsc{Verify}(\hat{\boldsymbol{y}}_g, \pi_{\text{target}})$
    \STATE Compute
    \[
    \Delta_g \leftarrow
    \sum_{t=1}^{K}
    \Big[
    \log \pi_{\text{target}}(y_t \mid \boldsymbol{x}, \boldsymbol{y}_{<t})
    -
    \log \pi_{\text{target}}(\hat{y}_{g,t} \mid \boldsymbol{x}, \hat{\boldsymbol{y}}_{g,<t})
    \Big]
    \]
    \STATE $r_g \leftarrow \dfrac{k_g}{k_g\gamma + 1} + \eta \cdot \mathbf{1}[\Delta_g<\epsilon]$
\ENDFOR
\STATE Compute normalized group-relative advantages $\{\hat{A}_g\}_{g=1}^{G_{\mathrm{roll}}}$ from rewards $\{r_g\}_{g=1}^{G_{\mathrm{roll}}}$

\STATE {\color{green!50!black}$\triangleright$ \textit{Policy update}}
\STATE Update $\theta$ using the clipped group-relative policy objective with target-anchored KL regularization

\ENDFOR
\end{algorithmic}
\end{algorithm}

\section{Analysis of Confidence-Weighted Draft--Target Divergence}
\label{app:proofs}

In Section~\ref{sec:adaptive_sampling}, we define the token-level \emph{criticality score}
\[
v_t = C(P_t)\, D_{\mathrm{KL}}(P_t \,\|\, Q_t),
\]
where $P_t$ and $Q_t$ denote the target and drafter distributions at position $t$, and $C(P_t)$ is an entropy-normalized confidence score of the target distribution. In the main text, this score serves as a prioritization signal for training windows. This appendix provides analytical support for this design by relating draft--target divergence to speculative acceptance behavior and clarifying the role of confidence weighting.

\subsection{Acceptance Probability and Draft--Target Divergence}

Let $\alpha_t$ denote the expected acceptance probability at position $t$ under speculative decoding. Following~\cite{leviathan2023fast}, when a draft token is sampled from $Q_t$ and verified against $P_t$, the expected acceptance probability is
\begin{align*}
\alpha_t 
&=
\mathbb{E}_{y\sim Q_t}\!\left[\min\!\left(1,\frac{P_t(y)}{Q_t(y)}\right)\right] \\
&=
\sum_{y\in\mathcal{V}} \min(P_t(y),Q_t(y)).
\end{align*}

Using the identity $\min(a,b)=\frac{1}{2}(a+b-|a-b|)$, we obtain
\begin{align*}
\alpha_t
&=
\frac{1}{2}\sum_{y\in\mathcal{V}}
\left(P_t(y)+Q_t(y)-|P_t(y)-Q_t(y)|\right) \\
&=
1-\frac{1}{2}\sum_{y\in\mathcal{V}}|P_t(y)-Q_t(y)| \\
&=
1-\delta(P_t,Q_t),
\end{align*}
where $\delta(P_t,Q_t)$ denotes the total variation distance.

Applying Pinsker's inequality,
\[
\delta(P_t,Q_t)
\le
\sqrt{\frac{1}{2}D_{\mathrm{KL}}(P_t \,\|\, Q_t)},
\]
yields
\[
\alpha_t
\ge
1-\sqrt{\frac{1}{2}D_{\mathrm{KL}}(P_t \,\|\, Q_t)}.
\]

This bound shows that smaller draft--target KL divergence implies a higher lower bound on the acceptance probability. It therefore supports the use of draft--target divergence as a proxy of speculative difficulty.

\subsection{Confidence Modulation of Draft--Target Divergence}

The analysis above shows that draft--target divergence is informative about acceptance behavior. However, the practical significance of a given divergence also depends on the uncertainty of the target distribution. For the same level of draft--target divergence, discrepancies are typically more consequential when the target distribution is concentrated than when the target itself is uncertain.

To account for this effect, we introduce the confidence factor
\[
C(P_t)=1-\frac{H(P_t)}{\log |\mathcal{V}|}\in[0,1],
\]
which assigns larger values to lower-entropy target distributions and smaller values to higher-entropy ones.

The resulting criticality score,
\[
v_t=C(P_t)\,D_{\mathrm{KL}}(P_t \,\|\, Q_t),
\]
can be viewed as a confidence-modulated measure of draft--target divergence. It increases the priority of positions where the drafter distribution diverges from a confident target distribution, while reducing the influence of divergence in higher-entropy contexts. Aggregating $v_t$ over a speculative window therefore yields a window-level signal that emphasizes regions more likely to constrain speculative acceptance in practice.

\section{Optimization Differences Between PPOW and Continued Supervised Training}
\label{app:sft_vs_rl}

PPOW and continued supervised training (CST) target different optimization objectives in speculative decoding. CST improves token-level imitation of the target model, whereas PPOW directly optimizes the window-level utility aligned with speculative decoding efficiency at inference time.

\subsection{Speculative Verification and Window-Level Utility}
\label{app:reject_sampling}

Speculative decoding commonly uses rejection-sampling verification to preserve the target-model distribution during inference. Under this verification rule, the practical gain of a speculative window depends on how many of its tokens are accepted by the target model, making the accepted prefix length, rather than token-level likelihood alone, the relevant measure of speculative utility.

For a speculative window $\hat{\boldsymbol{y}}=(\hat{y}_1,\dots,\hat{y}_K)$, the acceptance probability of the token at position $t$ under rejection-sampling verification is
\[
\alpha_t
=
\min\left(
1,\,
\frac{\pi_{\text{target}}(\hat{y}_{t}\mid \boldsymbol{x}, \hat{\boldsymbol{y}}_{<t})}
{\pi_{\theta}(\hat{y}_{t}\mid \boldsymbol{x}, \hat{\boldsymbol{y}}_{<t})}
\right).
\]
The acceptance length $\tau$ is then determined by the first position at which verification fails:
\[
\tau = \sum_{n=1}^{K}\prod_{t=1}^{n}\mathbf{1}[u_t \le \alpha_t],
\qquad
u_t \sim U(0,1).
\]
Because $\tau$ is defined by consecutive acceptance from the beginning of the speculative window, early truncation under verification immediately eliminates the contribution of all subsequent tokens in that window. In this sense, speculative utility is inherently window-level, and positions that cause early verification truncation can become bottlenecks with disproportionate impact on inference-time efficiency. This creates a mismatch with token-level training objectives and motivates the window-level optimization used in PPOW.

\subsection{Limitations of Continued Supervised Training (CST) for Speculative Decoding}

CST retains the same token-level cross-entropy objective as standard supervised training, and this token-level training objective is not fully aligned with the window-level speculative utility relevant at inference time:
\begin{align*}
\mathcal{L}_{\text{CST}}
&=
\mathbb{E}_{\boldsymbol{y}\sim \pi_{\text{target}}}
\left[-\log \pi_{\theta}(\boldsymbol{y}\mid \boldsymbol{x})\right] \\
&=
\mathbb{E}_{\boldsymbol{y}\sim \pi_{\text{target}}}
\left[\sum_{t=1}^{K}-\log \pi_{\theta}(y_t \mid \boldsymbol{x},\boldsymbol{y}_{<t})\right].
\end{align*}
This objective improves draft--target imitation. However, its effect on speculative decoding is indirect, because the optimized quantity is token-level likelihood rather than the realized utility of a speculative window. In other words, matching the target model token by token does not necessarily maximize the number of tokens that survive verification within an accepted speculative window.

As a result, CST does not directly optimize accepted prefix length, explicitly focus on bottleneck positions that cause early verification truncation, or optimize across multiple sampled speculative windows for the same context. These mismatches help explain why CST may yield limited additional gains in inference-time speculative efficiency, especially once token-level imitation has largely saturated, and motivate the window-level optimization used in PPOW.

\subsection{PPOW as Window-Level Policy Optimization}
PPOW formulates speculative decoding as a policy optimization problem, where the drafter is optimized over speculative windows using window-level rewards aligned with inference-time speculative performance (Section~\ref{sec:performance_rewards}):
\[
J(\theta)=\mathbb{E}_{\hat{\boldsymbol{y}}\sim \pi_{\theta}}
\big[R(\boldsymbol{x},\hat{\boldsymbol{y}})\big].
\]

Here, optimization is driven by the realized reward of sampled speculative windows rather than by token-level supervision on a fixed target token sequence. Since a full response can be decomposed into many overlapping speculative windows, not all windows contribute equally informative training signals. Window-level optimization therefore makes it possible to focus learning on more informative speculative windows, especially those containing positions that are more critical to early truncation under verification.

PPOW also samples multiple speculative windows for the same context and computes group-relative advantages among them. This grouped optimization reflects practical settings that evaluate multiple candidate token sequences during decoding and provides a stronger relative learning signal toward speculative windows with better acceptance behavior.

\section{Cost-Aware Speedup Reward versus Measured Speedup Reward}
\label{app:speedup_proxy}

A natural alternative to the Cost-Aware Speedup Reward in Section~\ref{sec:performance_rewards} is to define training rewards directly from measured inference speedup. In practice, however, measured speedup depends strongly on the serving environment, including hardware, backend implementation, cache behavior, batching, and verification configuration. It can also provide noisy or low-resolution feedback at the speculative-window level. This appendix compares the Cost-Aware Speedup Reward and a Measured-Speedup-Based Reward alternative from two perspectives: whether they exhibit similar reward trends as accepted length increases, and whether training with measured speedup yields meaningful gains in a fixed setup.

\begin{table}[htbp]
\centering
\caption{\textbf{Reward values induced by accepted prefix length.} Comparison between a Measured-Speedup-Based Reward and PPOW's Cost-Aware Speedup Reward as accepted length increases. The two rewards are not numerically identical, but they preserve the same monotonic trend.}
\label{tab:proxy_vs_measured}
\small
\begin{tabular*}{0.80\columnwidth}
{@{\hspace{5pt}\extracolsep{\fill}}ccc@{\hspace{5pt}}}
\toprule
Accepted Prefix Length & Measured Speedup Reward & Cost-Aware Speedup Reward \\
\midrule
1 & 0.81 & 0.89 \\[1pt]
2 & 1.49 & 1.60 \\[1pt]
3 & 1.81 & 2.18 \\[1pt]
4 & 2.06 & 2.67 \\[1pt]
5 & 2.49 & 3.08 \\[1pt]
6 & 2.62 & 3.43 \\[1pt]
7 & 2.88 & 3.74 \\[1pt]
\bottomrule
\end{tabular*}
\end{table}

\begin{table}[htbp]
\centering
\caption{\textbf{Training with PPOW's Cost-Aware Speedup Reward vs. Measured-Speedup-Based Reward.} Final average acceptance length ($\tau$) on GSM8K with LLaMA-3.1-8B under the same training budget.}
\label{tab:measured_reward_compare}
\small
\begin{tabular*}{0.6\columnwidth}{@{\hspace{5pt}\extracolsep{\fill}}lc@{\hspace{5pt}}}
\toprule
Reward formulation & Average acceptance length ($\tau$) \\
\midrule
Cost-Aware Speedup & 6.50 \\[1pt]
Measured Speedup & 6.61 \\[1pt]
\bottomrule
\end{tabular*}
\end{table}

Table~\ref{tab:proxy_vs_measured} compares the reward values induced by accepted prefix length under the two formulations. Although the Cost-Aware Speedup Reward does not numerically match the Measured-Speedup-Based Reward, the two exhibit the same monotonic trend: longer accepted prefixes receive larger reward under both formulations. The two rewards differ in scale because the measured reward is system-dependent, but they preserve the same ordering over accepted-prefix outcomes. This shared trend is important for training, since it preserves the relative preference for speculative windows with better efficiency characteristics.

Table~\ref{tab:measured_reward_compare} compares training with the Cost-Aware Speedup Reward against training with a Measured-Speedup-Based Reward under the same budget. In this fixed setup, the measured-speedup-based reward yields only a small improvement in final acceptance length. However, such a reward remains tightly coupled to the specific serving stack used during training and may provide noisy or low-resolution feedback at the window level. The cost-aware formulation therefore offers a more portable and practical default objective.

Overall, these results support the use of the Cost-Aware Speedup Reward during training: it tracks the same acceptance-related trend as measured speedup and remains effective for optimization, while avoiding the system dependence of direct runtime-based rewards.

\section{Effect of the Distribution-Based Proximity Reward Across Easy and Hard Windows}
\label{app:easy_hard_case}

To characterize the effect of the Distribution-Based Proximity Reward $R_{\text{dist}}$, we partition speculative windows according to the acceptance behavior of the supervised baseline. We refer to windows with full acceptance ($k=K$) as \emph{easy}, and to those for which verification terminates before the end of the speculative window ($k<K$) as \emph{hard}.

Under this partition, we compare the EAGLE-3 supervised baseline, PPOW without $R_{\text{dist}}$, and full PPOW. Table~\ref{tab:easy_hard_case} reports both average acceptance length $\tau$ and an alignment divergence metric $\nabla$, defined token-wise as
\[
\nabla = \exp(\delta)-\delta-1,
\qquad
\delta=\log \pi_{\text{target}}(\hat{y}_t\mid \cdot)-\log \pi_{\theta}(\hat{y}_t\mid \cdot).
\]
This metric measures draft--target alignment on drafted tokens through token-level log-probability differences, with lower values indicating better alignment.

\begin{table}[htbp]
\centering
\caption{\textbf{Effect of the Distribution-Based Proximity Reward on easy and hard windows.} Easy windows are fully accepted by the supervised baseline ($k=K$), whereas hard windows terminate early within the speculative window ($k<K$). Adding $R_{\text{dist}}$ preserves performance on easy windows and further improves acceptance length while reducing the alignment divergence metric $\nabla$ on hard windows.}
\label{tab:easy_hard_case}
\small
\setlength{\tabcolsep}{10pt}
\renewcommand{\arraystretch}{1.1}
\begin{tabular*}{0.80\columnwidth}
{@{\hspace{5pt}\extracolsep{\fill}}llcc@{\hspace{5pt}}}
\toprule
Window type & Model & $\tau$ & $\nabla$ \\
\midrule
\multirow{3}{*}{Easy} & EAGLE-3 & 6.0 & \textbf{0.009} \\
& PPOW (w/o $R_{\text{dist}}$) & 6.0 & 0.012 \\
& PPOW (Full)& 6.0 & 0.017 \\
\midrule
\multirow{3}{*}{Hard} & EAGLE-3 & 2.32 & 7.091 \\
& PPOW (w/o $R_{\text{dist}}$) & 4.58 & 3.644 \\
& PPOW (Full) & 4.86 & \textbf{3.275} \\
\bottomrule
\end{tabular*}
\end{table}

On easy windows, the supervised baseline, PPOW without $R_{\text{dist}}$, and full PPOW have identical acceptance length, and their $\nabla$ values differ only slightly. On hard windows, however, full PPOW improves acceptance length (4.58 $\rightarrow$ 4.86) while further reducing $\nabla$ (3.644 $\rightarrow$ 3.275) relative to PPOW without $R_{\text{dist}}$. These results support the inclusion of $R_{\text{dist}}$, which preserves performance on easy windows while yielding additional gains on hard windows.

\section{Sensitivity to the Weight ($\eta$) of the Distribution-Based Proximity Reward}
\label{app:eta_sensitivity}

The Distribution-Based Proximity Reward $R_{\text{dist}}$ is only applied when the accepted length is $k=0$, where the accepted-prefix reward alone provides no signal about whether the drafted window remains broadly compatible with the target model’s preferences. $R_{\text{dist}}$ therefore serves as a bounded auxiliary reward that provides partial credit in this sparse-reward case.

The coefficient $\eta$ controls the strength of this auxiliary reward relative to the main accepted-prefix / cost-aware speedup objective. In Table~\ref{tab:ablation}, the settings ``w/o $R_{\text{dist}}$'' and ``PPOW (Full)'' correspond to $\eta=0$ and $\eta=1.0$, respectively. To further study sensitivity to $\eta$, we additionally train models with $\eta=0.5$ and $\eta=2.0$ under the same LLaMA-3.1-8B setup and evaluate them on MT-Bench and GSM8K.

Table~\ref{tab:eta_sensitivity} shows that PPOW is reasonably stable when $\eta$ remains in a moderate range: $\eta=0.5$ and $\eta=1.0$ achieve similar results on both benchmarks. However, performance degrades at $\eta=2.0$, indicating that assigning too much weight to $R_{\text{dist}}$ can overemphasize distributional proximity and weaken optimization toward the main accepted-prefix / cost-aware speedup objective. Overall, these results suggest that $\eta$ should remain moderate rather than too large.

\begin{table}[h]
\centering
\caption{\textbf{Sensitivity to the weight $\eta$ of $R_{\text{dist}}$.} Results are reported under the same LLaMA-3.1-8B setup as Table~\ref{tab:ablation}.}
\label{tab:eta_sensitivity}
\small
\setlength{\tabcolsep}{2pt}
\renewcommand{\arraystretch}{1.25}
\begin{tabular*}{0.80\columnwidth}{@{\hspace{5pt}\extracolsep{\fill}}lcccc@{\hspace{5pt}}}
\toprule
\multirow{2}{*}{$\eta$} & \multicolumn{2}{c}{MT-Bench} & \multicolumn{2}{c}{GSM8K} \\[-3pt]
\cmidrule(lr){2-3} \cmidrule(lr){4-5}
& $\tau$ & Speedup & $\tau$ & Speedup \\[-2pt]
\midrule
0.0 & 5.05 & 2.52$\times$ & 6.41 & 3.48$\times$ \\
0.5 & 5.33 & 2.63$\times$ & 6.47 & 3.50$\times$ \\
\textbf{1.0} & \textbf{5.47} & \textbf{2.72}$\times$ & \textbf{6.50} & \textbf{3.52}$\times$ \\
2.0 & 5.18 & 2.57$\times$ & 6.32 & 3.36$\times$ \\
\bottomrule
\end{tabular*}
\end{table}

\section{Additional Baseline Comparisons}
\label{app:more_baseline}

\subsection{Comparisons with OSD and FastDraft}
\label{app:osd_fastdraft}

Table~\ref{tab:osd_fastdraft} reports additional comparisons against OSD~\cite{liu2024online} and FastDraft~\cite{zafrir2025fastdraft}. In the reported comparison, PPOW substantially improves over OSD on Vicuna-7B~\cite{zheng2023judging} across both GSM8K~\cite{cobbe2021training} and HumanEval~\cite{chen2021evaluating}. Compared with FastDraft on LLaMA-3.1-8B-Instruct, PPOW shows much larger gains on GSM8K and remains competitive on HumanEval, achieving a slightly higher acceptance length with comparable speedup. These results suggest that PPOW remains effective beyond the primary baselines reported in the main text.

\begin{table}[htbp]
\centering
\caption{\textbf{Additional baseline comparisons against OSD and FastDraft.} PPOW substantially outperforms OSD on Vicuna-7B and consistently improves over FastDraft on GSM8K, while remaining competitive on HumanEval with LLaMA-3.1-8B-Instruct.}
\label{tab:osd_fastdraft}
\small
\setlength{\tabcolsep}{10pt}
\renewcommand{\arraystretch}{1.1}
\begin{tabular*}{0.80\columnwidth}
{@{\hspace{5pt}\extracolsep{\fill}}lllcc@{\hspace{5pt}}}
\toprule
Target Model & Method & Benchmark & $\tau$ & Speedup\\
\midrule
\multirow{4}{*}{Vicuna-7B} & OSD & GSM8K & 3.53 & 1.35$\times$ \\
& PPOW & GSM8K & \textbf{6.12} & \textbf{4.40$\times$} \\
& OSD & HumanEval & 2.57 & 1.58$\times$ \\
& PPOW & HumanEval & \textbf{6.75} & \textbf{4.87$\times$} \\
\midrule
\multirow{4}{*}{LLaMA-3.1-8B-Instruct} & FastDraft & GSM8K & 2.67 & 1.63$\times$ \\
& PPOW & GSM8K & \textbf{6.50} & \textbf{3.52$\times$} \\
& FastDraft & HumanEval & 6.99 & \textbf{4.29$\times$} \\
& PPOW & HumanEval & \textbf{7.23} & 4.14$\times$ \\
\bottomrule
\end{tabular*}
\end{table}

\subsection{Additional Results on Natural-Language Tasks}
\label{app:nl_tasks}

Table~\ref{tab:nl_tasks} reports additional results on X-SUM~\cite{narayan2018don} and WMT14~\cite{bojar2014findings}. Compared with the stronger gains observed on reasoning tasks such as GSM8K, PPOW shows only modest improvements on X-SUM and limited gains on WMT14. We attribute this gap to the task properties of natural-language generation: these tasks admit a much more diverse set of valid token sequences, which weakens the acceptance-aware reward signal used by PPOW. The effect is less severe on X-SUM, where summaries remain partially anchored to source content, but becomes more pronounced on WMT14, where lexical and syntactic diversity greatly enlarges the space of valid outputs. These observations are consistent with the broader pattern that PPOW is most beneficial when speculative performance is controlled by relatively structured bottleneck decisions.

\begin{table}[htbp]
\centering
\caption{\textbf{Additional results on natural-language tasks.} PPOW slightly improves acceptance length and speedup on X-SUM, while gains on WMT14 are limited.}
\label{tab:nl_tasks}
\small
\setlength{\tabcolsep}{4pt}
\renewcommand{\arraystretch}{1.0}
\begin{tabular*}{0.80\columnwidth}{@{\extracolsep{\fill}}l cc cc cc}
\toprule
& \multicolumn{2}{c}{X-SUM} & \multicolumn{2}{c}{WMT14} & \multicolumn{2}{c}{Mean} \\
\cmidrule(lr){2-3} \cmidrule(lr){4-5} \cmidrule(lr){6-7}
Method & $\tau$ & Speedup & $\tau$ & Speedup & $\tau$ & Speedup \\
\midrule
\multicolumn{7}{c}{Temperature = 0.0} \\
\midrule
EAGLE3 & 4.94 & 2.81$\times$ & 2.72 & \textbf{1.56$\times$} & 3.83 & \textbf{2.19$\times$} \\
EAGLE3 + CST & 4.58 & 2.60$\times$ & \textbf{2.73} & \textbf{1.56$\times$} & 3.66 & 2.08$\times$ \\
PPOW & \textbf{5.13} & \textbf{2.87$\times$} & 2.61 & 1.49$\times$ & \textbf{3.87} & 2.18$\times$ \\
\midrule
\multicolumn{7}{c}{Temperature = 1.0} \\
\midrule
EAGLE3 & 4.15 & 2.30$\times$ & \textbf{2.56} & \textbf{1.44$\times$} & 3.36 & 1.87$\times$ \\
EAGLE3 + CST & 3.87 & 2.17$\times$ & 2.50 & 1.37$\times$ & 3.19 & 1.77$\times$ \\
PPOW & \textbf{4.44} & \textbf{2.53$\times$} & 2.37 & 1.30$\times$ & \textbf{3.41} & \textbf{1.92$\times$} \\
\bottomrule
\end{tabular*}
\end{table}

\section{Limitations and Broader Impacts}
\label{app:limitations and impacts}

\subsection{Limitations}
\label{app:limitations}
PPOW improves speculative decoding by aligning training with inference-time acceptance behavior, but this stronger alignment comes with additional training overhead. Compared with supervised drafter training, PPOW requires grouped rollouts, speculative verification, and target-anchored KL regularization, and our current implementation uses both a frozen target model and a trainable drafter. This makes PPOW more resource-intensive and more complex to implement than continued supervised training.

This alignment objective also introduces several hyperparameters beyond standard reinforcement learning, including speculative-decoding-specific settings such as the window size, the relative cost $\gamma$, the proximity reward weight $\eta$, the proximity threshold $\epsilon$, and the hard-window curriculum in ADAW. While we use a unified configuration across experiments and observe stable gains, reducing such dependence through more adaptive or self-tuning variants is a natural direction for future work.

\subsection{Broader Impacts}
\label{app:impacts}
PPOW suggests an additional algorithmic perspective on LLM inference: some components of the inference pipeline may be optimized directly with performance-driven objectives rather than only with token-level supervision. In this setting, reinforcement learning provides a practical way to align training with system-level inference behavior when the utility of a model component is determined by structured interactions during decoding rather than by local prediction quality alone.

Framing speculative decoding as a learning problem may also relate to other inference-time decisions, such as candidate allocation, request routing, scheduling, and load balancing. These components are often optimized separately, but in practice they jointly affect end-to-end serving efficiency. The formulation in PPOW may offer a useful perspective for future work that treats such inference-time modules as part of a unified decision-making environment.

\end{document}